%% file: AISTATS_main.tex
\documentclass[twoside]{article}

\usepackage[accepted]{aistats2025}

\input{notation}
\begin{document}

\twocolumn[

\aistatstitle{Provable Benefits of Task-Specific Prompts for In-context Learning}
\aistatsauthor{Xiangyu Chang$^1$ \and Yingcong Li$^2$ \and Muti Kara$^3$ \and Samet Oymak$^2$\and Amit K. Roy-Chowdhury$^1$}
\aistatsaddress{$^1$University of California, Riverside \and $^2$University of Michigan \and $^3$Bilkent University}]

\begin{abstract}

The in-context learning capabilities of modern language models have motivated a deeper mathematical understanding of sequence models. A line of recent work has shown that linear attention models can emulate projected gradient descent iterations to implicitly learn the task vector from the data provided in the context window. In this work, we consider a novel setting where the global task distribution can be partitioned into a union of conditional task distributions. We then examine the use of task-specific prompts and prediction heads for learning the prior information associated with the conditional task distribution using a one-layer attention model. Our results on loss landscape show that task-specific prompts facilitate a \emph{covariance-mean decoupling} where prompt-tuning explains the conditional mean of the distribution whereas the variance is learned/explained through in-context learning. Incorporating task-specific head further aids this process by entirely decoupling estimation of mean and variance components. This covariance-mean perspective similarly explains how jointly training prompt and attention weights can provably help over fine-tuning after pretraining. The code for reproducing the numerical results is available at \href{https://github.com/xchang1121/prompt_ICL}{GitHub.}

\end{abstract}

\input{sec/intro}
\input{sec/related}
\input{sec/setup}

\input{sec/main_v2}

\input{sec/vector_head}
\input{sec/exp}

\input{sec/acknowledgement}


\bibliography{ref}
\bibliographystyle{plainnat}

\input{app/checklist}

\newpage
\appendix
\onecolumn
\aistatstitle{Appendix}
\input{AISTATS_supp}

\end{document}

%% file: notation.tex
\usepackage[utf8]{inputenc} 
\usepackage{natbib}
\usepackage[T1]{fontenc}    
\usepackage{titletoc}
\usepackage{hyperref}       
\usepackage{url}            
\usepackage{booktabs}       
\usepackage{amsfonts,amsmath,amssymb}       
\usepackage{nicefrac}       
\usepackage{microtype}      
\usepackage{xcolor}         
\usepackage{enumitem}
\usepackage{mathtools}

\usepackage{graphicx}
\usepackage{subcaption}
\usepackage[normalem]{ulem}
\usepackage[textwidth=1.8cm]{todonotes}
\usepackage{ragged2e}
\usepackage{minitoc}
\usepackage{wrapfig,bm,comment,color}
\usepackage{breakurl,epsfig,epsf,fmtcount,semtrans,multirow,multicol,boldline}
\usepackage{tcolorbox}
\tcbuselibrary{skins}
\usepackage{tikz}
\usepackage{pifont}

\usepackage{appendix}
\usepackage{xcolor}

\definecolor{darkred}{RGB}{150,0,0}
\definecolor{darkgreen}{RGB}{0,150,0}
\definecolor{darkblue}{RGB}{0,0,200}
\hypersetup{colorlinks=true, linkcolor=darkred, citecolor=darkgreen, urlcolor=darkblue}

\newtheorem{theorem}{Theorem}

\newtheorem{assumption}{Assumption}

\newtheorem{lemma}{Lemma}
\newtheorem{corollary}{Corollary}
\newtheorem{proposition}{Proposition}
\newtheorem{definition}{Definition}

\newcommand{\beq}{\begin{equation}}
\newcommand{\ba}{\begin{align}}
\newcommand{\ea}{\end{align}}

\newcommand{\eeq}{\end{equation}}

\def \endprf{\hfill {\vrule height6pt width6pt depth0pt}\medskip}

\newenvironment{proof}{\noindent {\bf Proof.} }{\endprf\par}

\newcommand{\vct}[1]{\bm{#1}}
\newcommand{\tr}[1]{\texttt{tr}(#1)}
\newcommand{\mtx}[1]{\bm{#1}}
\newcommand{\red}{\textcolor{darkred}}

\newcommand{\chang}[1]{\textcolor{red}{[Chang: #1]}}

\newcommand{\eps}{\varepsilon}

\newcommand{\hb}{\vct{h}}
\newcommand{\hbb}{\bar{\vct{h}}}

\newcommand{\st}{\star}

\newcommand{\A}{{\mtx{A}}}

\newcommand{\B}{{\mtx{B}}}

\newcommand{\Ib}{{{\mtx{I}}}}

\newcommand{\Lc}{{\cal{L}}}

\newcommand{\Nc}{{\cal{N}}}

\newcommand{\Dc}{{\cal{D}}}

\newcommand{\Pb}{{\mtx{P}}}

\newcommand{\Cb}{{\mtx{C}}}

\newcommand{\Hb}{{\mtx{H}}}

\newcommand{\bmu}{{\boldsymbol{\mu}}}

\newcommand{\onebb}{{\mathbf{1}}}
\newcommand{\zerbb}{{\mathbf{0}}}
\newcommand{\Iden}{{\mtx{I}}}
\newcommand{\M}{{\mtx{M}}}
\newcommand{\z}{{\vct{z}}}

\newcommand{\bt}{{\boldsymbol{\beta}}}

\newcommand{\btb}{\bar{\boldsymbol{\beta}}}

\newcommand{\vb}{\vct{v}}

\newcommand{\ft}{\textnormal{FT}}
\newcommand{\pt}{\textnormal{PT}}
\newcommand{\jt}{\textnormal{JT}}

\newcommand{\cb}{\mtx{c}}
\newcommand{\w}{\vct{w}}

\newcommand{\ab}{\vct{a}}

\newcommand{\Z}{\mtx{Z}}
\newcommand{\Zk}{\mtx{Z}^{(k)}}

\newcommand{\bSi}{\boldsymbol{\Sigma}}
\newcommand{\bSibt}{\boldsymbol{\Sigma}_{\bt}}

\newcommand{\bSbbt}{\boldsymbol{\bar\Sigma}_{\bt}}
\newcommand{\bStbt}{\boldsymbol{\tilde\Sigma}_{\bt}}
\newcommand{\bSbmu}{\boldsymbol{\bar\Sigma}_{\bmu}}
\newcommand{\bSb}{\boldsymbol{\bar{\Sigma}}}
\newcommand{\bSp}{\boldsymbol{\tilde{\Sigma}}}

\newcommand{\x}{\vct{x}}

\newcommand{\y}{\vct{y}}

\newcommand{\W}{\mtx{W}}
\newcommand{\WK}{\mtx{W}_k}
\newcommand{\WQ}{\mtx{W}_q}
\newcommand{\WV}{\mtx{W}_v}

\newcommand{\X}{{\mtx{X}}}

\newcommand{\p}{{\vct{p}}}

\newcommand{\pb}{{\vct{p}}}
\newcommand{\pbb}{{\bar{\vct{p}}}}

\newcommand{\R}{\mathbb{R}}

\newcommand{\E}{\operatorname{\mathbb{E}}}

\newcommand{\nn}{\nonumber}

\newcommand{\tn}[1]{\left\|{#1}\right\|_{\ell_2}}

\newcommand{\order}[1]{{\cal{O}}(#1)}

\newcommand{\Wt}{\tilde{\mtx{W}}}

\newcommand{\Wb}{\mtx{\bar{W}}}

\newcommand{\pgd}{{\textnormal{PGD}}}
\newcommand{\att}{{\textnormal{Attn}}}

\newcommand\scalemath[2]{\scalebox{#1}{\mbox{\ensuremath{\displaystyle #2}}}}

\newcommand\DoToC{%
  \startcontents
  \printcontents{}{1}{\textbf{Contents}\vskip3pt\hrule\vskip5pt}
  \vskip3pt\hrule\vskip5pt
}

\newcommand{\todoasr}[1]{} 
\newcommand{\ankit}[1]{}
\newcommand{\asrnote}[1]{}

\newcommand{\bias}[1]{{\textcolor{darkblue}{#1}}}
\newcommand{\debias}[1]{{\textcolor{darkgreen}{#1}}}

%% file: sec/intro.tex
\section{Introduction}

\input{AISTATS_figs/intro_fig}

Modern language models possess a remarkable ability to learn new tasks or solve complex problems using examples provided within their context window~\citep{brown2020language, team2023gemini, gpt4_techreport, touvron2023_llama}. This capability, known as \textit{in-context learning} (ICL), offers a novel and efficient alternative to traditional fine-tuning methods. ICL enables models to adapt to a wide range of tasks through a single forward pass, eliminating the need for task-specific weight updates. This adaptability has made ICL a central feature in the use of large language models (LLMs), extending their utility across diverse applications. 

In a basic ICL setting, we construct an input sequence $\Z$ that contains a query to label and related input-label demonstrations. We feed $\Z$ to a sequence model $f$ to predict the label $y$ of this query. Thus, the ICL optimization can be written as
\begin{align}
\min_{\W}\E_{y,\Z}[\text{loss}(y, f_{\W}(\Z))],\label{vanilla_icl}
\end{align}
where $\W$ denotes the weights of $f$. In practice, however, ICL examples are typically paired with task-specific prompts. These prompts can be manually crafted or automatically generated using methods like differentiable optimization or zero-order search. In fact, the model can often solve the task without any ICL example (zero-shot) by solely relying on the prompt \red{\cite{}}. This motivates a deeper understanding of the synergies between prompt-tuning and in-context learning. Concretely, consider a multitask learning setting where we first sample a task $t$ and then sample $(\Z,y)$ from the associated task distribution. In this case, a standard approach is crafting a dedicated prompt $\pb_t$ to feed with the input sequence. The joint optimization of the weights $\W$ and prompts $\Pb=(\pb_t)_{t\ge 1}$ takes the form
\begin{align}
\min_{\W,\Pb}\E[\text{loss}(y, f_{\W}(\Z,\pb_t))].\label{joint_icl}
\end{align}
Contrasting \eqref{vanilla_icl} with \eqref{joint_icl} motivate a few fundamental questions: 
\begin{enumerate}[label=Q\arabic*.]
\item How do ICL and prompt-tuning synergistically contribute to learning?
\item In practice, we first pretrain a model via \eqref{vanilla_icl} and then tune a task-specific prompt. Does joint training have an advantage over this?
\item Can utilizing additional task-specific parameters together with prompts, further boost performance?
\end{enumerate}

To answer these questions, we conduct a comprehensive investigation of the optimization landscape for attention weights and task-specific prompts within a multitask dataset model, examining both joint optimization approaches and sequential strategies involving attention weight pretraining followed by prompt-tuning. We derive closed-form solutions for optimal parameters and loss landscapes, introducing the novel concept of "covariance-mean decoupling" to elucidate the impact of different training strategies on model performance. While previous theoretical research has primarily focused on attention weight optimization, we extend the analysis to include tunable prompts in a multi-task linear regression framework, addressing the practical scenario where models fine-tune task-specific modules on fixed backbones. Notably, our work advances beyond existing research by incorporating non-zero task mean and non-isotropic covariance considerations, revealing why fine-tuning may not consistently enhance performance. Our theoretical framework demonstrates how varying performance gains across training strategies can be attributed to covariance-mean decoupling, providing both theoretical foundations and practical insights for optimizing attention-based models through careful design of tunable components.

Our key contributions are:

\begin{enumerate}
\item \textbf{Comprehensive analysis of training strategies:} We analyze multi-task linear regression with a linear attention layer and tunable parameters, covering joint optimization and pretraining-finetuning methods. A unified parameterization allows for closed-form solutions of optimal parameters and loss landscapes, generalizing prior work. (see Section~\ref{sec:main})

\item \textbf{Mean-covariance decoupling concept:} We introduce covariance-mean decoupling through closed-form loss landscape analysis, demonstrating its correlation with model performance—the greater the decoupling, the better the model's predictions. (see Section~\ref{sec:decoupling})

\item \textbf{Path to optimal in-context learning:} Our analysis guides the design of attention models, emphasizing the importance of training sequence and parameter selection. We propose a model design achieving full covariance-mean decoupling, offering a strategy for improving in-context learning. (see Theorem~\ref{thm:fully_decoupled})
\end{enumerate}

These contributions provide a deeper understanding of prompt-tuning and weight optimization, offering insights for designing more effective models that minimize loss and maximize performance.

%% file: AISTATS_figs/intro_fig.tex
\begin{figure}[!t]
\includegraphics[width=0.5\textwidth]{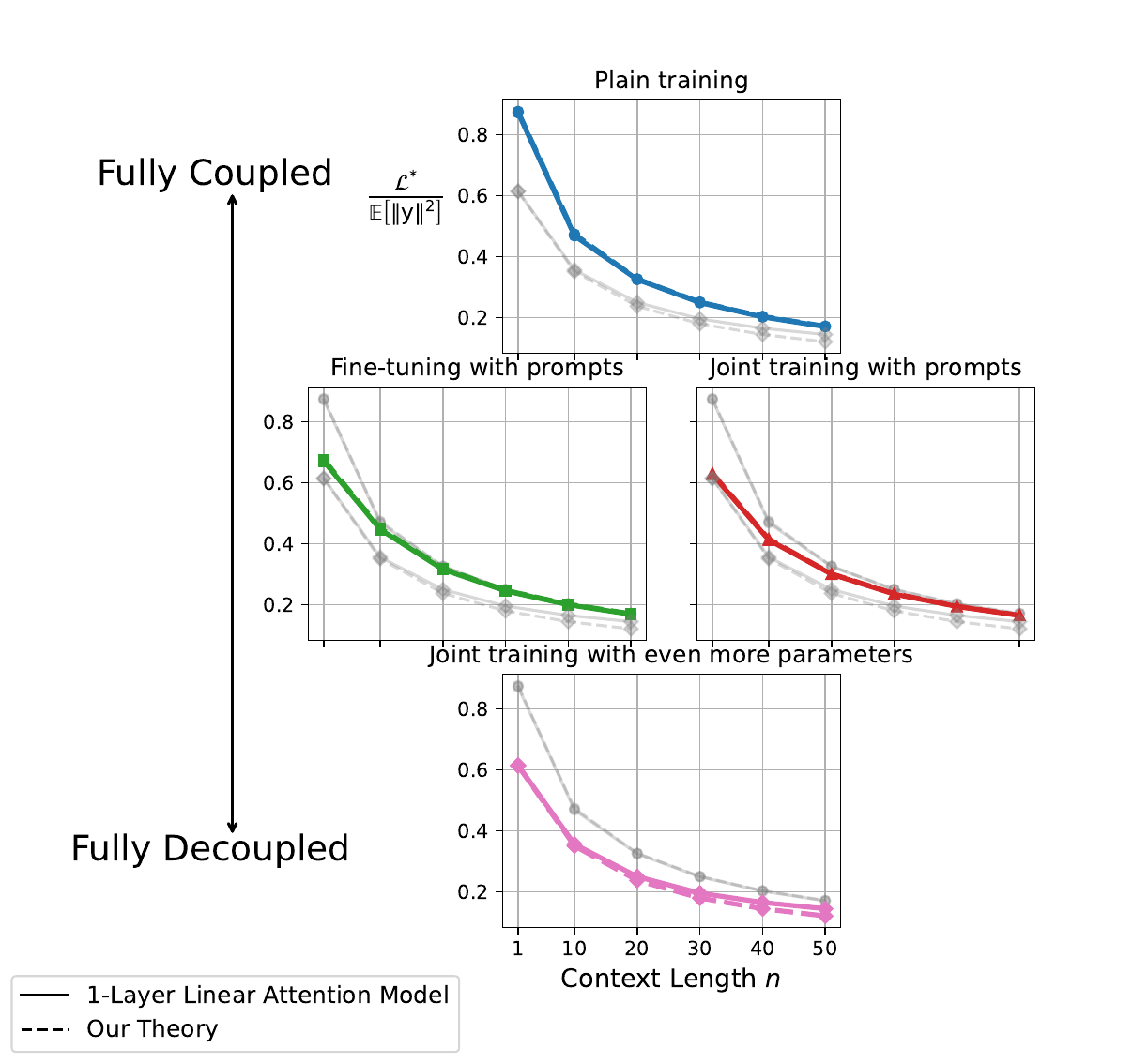}
\caption{Overview of our work: We present a theoretical analysis of a 1-layer linear attention model for multi-task in-context learning (ICL), examining different configurations of task-specific parameters. By progressively introducing more task-specific parameters across various training settings, we achieve complete \textbf{covariance-mean decoupling}, leading to an optimal multi-task ICL loss. In our analysis, we derive upper and lower bounds for the multi-task ICL loss, corresponding to fully coupled and fully decoupled covariance-mean scenarios.}
\end{figure}

%% file: sec/related.tex
\section{Related work}

Understanding in-context learning (ICL) in large language models (LLMs) has become a key research focus~\citep{brown2020language, liu2023pre, rae2021scaling}, particularly due to LLMs' ability to generalize across diverse applications~\citep{team2023gemini, gpt4_techreport, touvron2023_llama}. ICL enables models to adapt to new tasks using examples provided during inference without parameter updates, effectively serving as meta-learners. This has led to research exploring how LLMs leverage in-context information.

Several studies have linked ICL with gradient-based learning mechanisms. \citet{akyrek2023what} and \citet{von2023transformers} show that Transformers can emulate gradient descent (GD) steps using in-context examples, suggesting that Transformers implicitly learn gradient-based updates within their attention mechanisms.

Recent work has provided theoretical perspectives on ICL in simpler models like single-layer linear attention. \citet{zhang2024trained, mahankali2024one, ahn2024transformers, li2023dissecting} examine how these models can emulate GD-like algorithms when trained on in-context prompts. \citet{mahankali2024one} and \citet{ahn2024transformers} demonstrate that models trained on isotropic Gaussian data perform GD steps at test time, while \citet{li2023transformers} explores generalization bounds for multi-layer Transformers. \citet{li2024fine} extends this to dependent data with single-task ICL, showing that centralized data enables optimal preconditioned GD steps. However, these works primarily focus on zero-mean distributions or single parameters (W), simplifying the optimization landscape.

Our work addresses multi-task ICL with non-zero means and varying covariances, expanding beyond zero-mean assumptions in prior studies~\citep{li2023transformers, li2024fine}. We explore the joint optimization of $\W, \pb$, and $\hb$ parameters, introducing task-specific structures that reduce the influence of task-specific means. Unlike \citet{li2024fine} who focus on single-task scenarios or \citet{li2023transformers} who assume a single mean and covariance, we analyze diverse task distributions with distinct means and covariances, developing the novel concept of "mean-covariance decoupling" to reveal how task-specific parameters enhance ICL performance. Our approach provides theoretical guarantees and empirical evidence for prompt-tuning in complex, multi-task environments with real-world non-zero mean distributions.

%% file: sec/setup.tex
\section{Setup and Preliminaries}
\label{sec:assumptions}
We begin with a brief note on notation. 
Let $[n]$ denote set $\{1,\cdots,n\}$ for some integer $n$.
Bold lowercase and uppercase letters (e.g., $\x$ and $\X$) represent vectors and matrices, respectively. 
$\onebb_d$ and $\zerbb_d$ refer to the $d$-dimensional all-ones and all-zeros vectors, respectively, while $\Iden_d$ denotes the $d \times d$ identity matrix. 
Additionally, $\tr{\W}$ represents the trace of the square matrix $\W$.

Our results are presented in a finite-dimensional setting.


\subsection{In-context learning} \label{sec:icl}

We consider an in-context learning (ICL) problem with demonstrations $(\x_i,y_i)_{i=1}^{n+1}$, and the input sequence $\Z$ is defined by removing $y_{n+1}$ as follows:
\begin{align}
\Z&=[\z_1~\dots~\z_n~\z]^\top=\begin{bmatrix}
    \x_1&\dots&\x_n&\x\\
    y_1&\dots&y_n&0
\end{bmatrix}^\top\notag\\
&=\begin{bmatrix}
    \X^\top&\x\\
    \y^\top&0
\end{bmatrix}^\top\in\R^{(n+1)\times(d+1)}.
\label{def Z}
\end{align}
 
 Here, $\z=[\x^\top~0]^\top$ is the query token where $\x:=\x_{n+1}$, and $\X=[\x_1~\cdots~\x_n]^\top\in\R^{n\times d}$, $\y=[y_1~\cdots~y_n]^\top\in\R^n$. 
 Then, we aim for a sequence model to predict the associated label  $y:=y_{n+1}$ of the given input sequence $\Z$. In this work, we consider the following data generation of $(\Z,y)$. We will refer to $(\X, \y)$, $\x$, and $y$ as contexts, query feature and the label to predict, respectively.
\begin{definition}[Single-task ICL]
\label{definition:single_task}
 Given a task mean $\bmu\in\R^d$, and covariances $\bSi_{\x},\bSi_{\bt}\succ 0\in\R^{d\times d}$. The input sequence and its associated label, i.e., $(\Z,y)$ with $\Z$ denoted in \eqref{def Z}, 
 are generated as follows:
\begin{itemize}
    \item A task parameter $\bt$ is generated from a Gaussian prior $\bt \sim \mathcal{N}(\bmu,\bSi_\bt)$.
    \item Conditioned on $\bt$, for $i\in[n+1]$, $(\x_i, y_i)$ is generated by $\x_i \sim \mathcal{N}(0, \bSi_{\x})$ and $y_i\sim \mathcal{N}(\x_i^\top\bt, \sigma^2)$.
\end{itemize}
\end{definition}
Here, $\sigma \geq 0$ is the noise level.

In this work, we study the task-mixture ICL problem where the task parameter $\bt$ of each input sequence is sampled from $K$ different distributions, $K\geq1$.

\begin{definition}
[Multi-task ICL]
\label{definition:multi-task}
Consider a multi-task ICL problem with $K$ different tasks. 
Each task generates $(\Z,y)\sim\Dc_k$ following Definition~\ref{definition:single_task} using shared feature distribution $\x_i \sim \mathcal{N}(0, \bSi_{\x})$, $i\in[n+1]$ but distinct task distributions 
$\bt_k \sim  \mathcal{N}(\bmu_k, \bSi_{\bt_k})$
with mean $\bmu_k$ and covariance $\bSi_{\bt_k}$ for $k \in [K]$.

Additionally, let $\{\pi_k\}_{k=1}^K$ be the probabilities of each task, satisfying $\sum_{k=1}^K \pi_k = 1$ and $\pi_k \geq 0$. 
\end{definition}

We consider a task-aware multi-task ICL setting. Specifically, when a task is selected according to $\pi_k$, its task index $k$ is known.

Let $\bar\Dc:=\sum_{k=1}^K\pi_k\Dc_k$ be the mixture of distributions and given sequence model $f:\R^{(n+1)\times(d+1)}\to\R$, we define the multi-task ICL objective as follows:
\begin{align}
\Lc(f)&=\E_{(\Z,y)\sim\bar\Dc}[\left(y-f(\Z)\right)^2].\label{obj}
\end{align}

Notably, the multi-task ICL defined in Definition~\ref{definition:multi-task} differs from conventional multi-task learning \citep{caruana1997multitask,zhang2021survey,li2023provable} where finite examples optimize task-specific parameters. We instead parameterize task distributions with $\bmu_k$ and $\bSi_{\bt_k}$, sampling unseen test tasks $\bt_k$ and focusing on distribution-level generalization. We address the meta-learning objective in \eqref{obj}, treating each distribution as a meta-learning problem (c.f.~Definition~\ref{definition:single_task}).

\subsection{Single-layer linear attention}
Considering the task-aware multi-task ICL problem as described in Section~\ref{sec:icl}, we explore the benefits of using \emph{task-specific prompts} to enhance the performance. 
\begin{definition}[Task-specific prompts]
\label{def:task-specific param} Recap input sequence $\Z$ from \eqref{def Z}. 
Given a task index $k$, $k\in[K]$, let $\pb_k\in\R^{d+1}$ represent its corresponding trainable prompt token. Then the input sequence of task $k$ is denoted by:
\begin{align}
    \Zk
    =[\pb_k~\z_1~\dots~\z_n~\z]^\top\in\R^{(n+2)\times (d+1)}.\label{def Z k}
\end{align}
\end{definition}

Our work focuses on the single-layer linear attention model in solving multi-task ICL problem with data distribution following Definition~\ref{definition:multi-task}.  
Given an input sequence $\Zk$ corresponding to task $k$ as defined in \eqref{def Z k}, let $\WQ, \WK, \WV \in \R^{(d+1) \times (d+1)}$ denote the query, key, and value parameters.
Then the single-layer linear attention model outputs
\begin{align*}
\att(\Zk)=(\Zk\WQ\WK^\top(\Zk)^\top)\M\Zk\WV
\end{align*}
where $\att(\cdot):\R^{ (n+2)\times (d+1)}\to \R^{ (n+2)\times (d+1)}$. 
Here, inspired by the previous work {\citep{ahn2024transformers}}, we apply mask matrix $\M=\begin{bmatrix}
    \Iden_{n+1}&\zerbb_{n+1}\\\zerbb_{n+1}^\top&0
\end{bmatrix}$ to to separate the $(n+1)$-colomun context and the query $\z$ in $\Zk$. Let $\hb\in\R^{d+1}$ be the linear head that enables the single-layer linear attention model to map the input sequence to the prediction. Additionally, for simplification and without loss of generality, let $\A:=\WQ\WK^\top$ and $\ab:=\WV\hb$. Then the prediction returns

\begin{align}
\hat y:=f_{\att}(\Zk)=(\z^\top\A(\Zk)^\top)\M\Zk\ab.\label{eqn:LinearAttn}
\end{align}

This is a more general form than the widely discussed \cite{wu2023many} case, which uses the {bottom-right} entry of the attention layer output as prediction (and equivalent to a one-hot head $\hb=\mtx{e}_{d+1}$), i.e., $f_{\att}(\Zk)=\att(\Zk)_{n+2,d+1}$.


\subsection{Optimizing the attention model}
\label{sec:PGD}
Our goal is to understand how optimizing $f_\att$ in \eqref{eqn:LinearAttn} results in in-context learning. 
To this aim, we introduce the following widely applied \cite{wu2023many,ahn2024transformers} assumption on the model construction. 
%
\begin{assumption}[Preconditioning]Given parameters $\A:=\WQ\WK^\top$ and $\ab:=\WV\hb$, they are constrained by
\[
\A = \begin{bmatrix} 
\W_{d\times d} & \zerbb_{d \times 1} \\
*_{1\times d} & *_{1\times 1} 
\end{bmatrix},~\\
\ab = \begin{bmatrix} 
 \zerbb_{d\times 1} \\
 1_{1\times 1} 
\end{bmatrix}.
\]
Here, we use $*$ to fill the entries that do not affect the final prediction, with subscripts indicating the dimensions.
$\W\in \R^{d\times d}$  represents the parameter that governs \(\A\). 
\label{assumption:PGD} 
\end{assumption}

Under Assumption~\ref{assumption:PGD}, let the prompt token for task $k$ be $\pb_k = \begin{bmatrix} \pbb_k \\ 1 \end{bmatrix}$, hence $\Zk=\begin{bmatrix}
    \pbb_k&\X^\top&\x\\
    1&\y^\top&0
\end{bmatrix}^\top$. The prediction of a single-layer linear attention model in \eqref{eqn:LinearAttn} can then be written as:
\begin{align}
f_{\att}(\Zk)& =\x^\top\W\begin{bmatrix}
    \bar{\pb}_k&\X^\top
\end{bmatrix}\begin{bmatrix}
    1\\\y
\end{bmatrix}\notag\\&=\x^\top \W (\X^\top \y + \pbb_k).
\label{eqn:pgd}
\end{align}
It is worth noting that we set the last entry of $\ab$ and each $\pb_k$ for $k \in [K]$ to one for simplicity, as any nonzero value yields the same output of~\eqref{eqn:pgd} due to rescaling invariance of:  $\ab \leftarrow \gamma_1 \ab$, $\A \leftarrow \gamma_1^{-1} \A$, and $\pbb_k \leftarrow \begin{bmatrix} \gamma_2^{-1} \pbb_k \\ \gamma_2 \end{bmatrix}$ for any nonzero scalar $\gamma_1,\gamma_2$.

We define the set of tunable prompts as:
\begin{align}
\Pb = [\pbb_1~\dots~\pbb_K]^\top \in \R^{K \times d}.\label{eqn:prompt}
\end{align}

Notably, previous research \cite{ahn2024transformers} has rigorously proven that for single-layer linear attention applied to a single-task linear regression ICL problem with zero-mean features, i.e., $\E[\x_i]=\E[\bt]=\zerbb$, the optimal solution must conform to the structure specified in Assumption~\ref{assumption:PGD}. This insight motivates our adoption of this assumption when extending the analysis to more complex multi-task ICL settings with non-zero task means. Building on this foundation, our work breaks new ground by exploring task-specific tuning for ICL across multiple tasks with varying means, offering a more general perspective on ICL. This represents a significant advancement in the field, as understanding the loss landscape in the simpler $\W$-preconditioned space is an essential step toward tackling the complexities of the full $\WK, \WQ, \WV$ parameter space.

Additionally, in Section~\ref{sec:fully-decoupled}, we show that it is possible to derive closed-form optimal solutions for $\WK, \WQ, \WV$ with both task-specific prompts and heads, even without relying on Assumption~\ref{assumption:PGD}, further expanding the scope of our analysis.



Recall the attention predictor from \eqref{eqn:pgd} and loss function from \eqref{obj}. Consider the multi-task ICL problem defined in Definition~\ref{definition:multi-task} and let $\W\in\R^{d\times d}$ and $\Pb\in\R^{K\times d}$ be the tunable parameters corresponding to attention weights and task-specific prompt tokens. The multi-task ICL loss is given by
\begin{align}
    &\Lc(\W,\Pb)=\sum_{k=1}^K \pi_k \Lc_k(\W,\pbb_k) \label{eqn:loss}\\
    &\text{where }\Lc_k(\W,\pbb_k)=\E_{(\Z,y)\sim\Dc_k}\left[( f_\att(\Z^{(k)})-y )^2\right].\nn
\end{align}

In this work, we address multi-task ICL problems by investigating and comparing three optimization settings: plain training, fine-tuning, and joint training.

\noindent\textbf{Plain training: }Plain training refers to a standard ICL problem that train an linear attention model without applying the task-specific prompts that are defined in Definition~\ref{def:task-specific param}. Therefore, following loss function \eqref{eqn:loss}, its objective can be defined via
\begin{align}
    \W^\st_\pt=\arg\min_{\W}\Lc(\W,\Pb=\zerbb),~\label{eqn:obj pt}
\end{align}
and $\Lc_\pt^\st=\Lc(\W^\st_\pt,\Pb=\zerbb)$ is the optimal loss.

\noindent\textbf{Fine-tuning: }Fine-tuning/Prompt-tuning involves training separate prompts for each task while keeping the attention weights fixed. 
The goal then is to fine-tune the prompt parameters $\Pb$ (c.f.~\eqref{eqn:prompt}) for all the tasks $k\in[K]$. Suppose that parameter $\W$ is given, the the optimal $\Pb$ based on $\W$ is defined by:
\begin{align*}
    \Pb^\st(\W)=\arg\min_{\Pb}\Lc(\W,\Pb).
\end{align*}
In this work, we consider fine-tuning based on the plain pretrained model, that is, by setting $\W=\W^\st_\pt$ given in \eqref{eqn:obj pt}, and define the optimal solution by
\begin{align}
    \Pb^\st_\ft:=\Pb^\st(\W^\st_\pt)=\arg\min_{\Pb}\Lc(\W^\st_\pt,\Pb).\label{eqn:obj ft}
\end{align}
The optimal loss is given via $\Lc_\ft^\st=\Lc(\W^\st_\pt,\Pb^\st_\ft)$.


\noindent\textbf{Joint training: }In contrast, joint training involves jointly optimizing the attention weights $\W$ and prompt tokens $\Pb$. 
Hence, the optimization problem can be formulated as:
\begin{align}
    \W^\st_\jt,\Pb^\st_\jt=\arg\min_{\W,\Pb}\Lc(\W,\Pb).\label{eqn:obj jt}
\end{align}
The optimal loss is given via  $\Lc_\jt^\st=\Lc(\W^\st_\jt,\Pb^\st_\jt)$.

%% file: sec/main_v2.tex
\section{Main Results}\label{sec:main}
  In this section, we train and optimize the single-layer linear attention model in a multi-task linear regression ICL setting with dataset described in Definition~\ref{definition:multi-task}, and characterize the loss landscape  under different settings, i.e., $\Lc^\st_\pt$, $\Lc^\st_\ft$ and $\Lc^\st_\jt$ in Section~\ref{sec:PGD}. 
\subsection{Optimization landscape  }
\label{sec:landscape}
Recap the multi-task ICL dataset from Definition \ref{definition:multi-task} where $\bSi_{\x}$ is the shared covariance matrix of the input features and $\{(\bmu_k,\bSi_{\bt_k})\}_{k=1}^K$ are the task mean vectors and covariance matrices. In the main paper, we consider noiseless data setting where $\sigma=0$. We defer the exact analysis considering noisy labels to Appendix.

Consider any data distribution in Definition~\ref{definition:single_task} and let $\bt\sim\Nc(\bmu,\bSi_\bt)$.   $\bt$ can be rewritten via  $\bt=\btb+\bmu$ with $\btb\sim\Nc(0,\bSi_{\bt})$.  Then under the noiseless setting ($\sigma=0$), the associated label $y_i=\x_i^\top\bt$ is generated via
\begin{align}
    y_{i}
    &={\underset{\text{debiased}}{\underbrace{\x_i^\top\btb}}+\x_i^\top\bmu.}\label{eqn:bias}
\end{align}
Here we describe $\x_i^\top\btb$ as debiased since $\E[\x_i^\top\btb]=0$. 
In this work, we investigate how task-specific prompts can help to capture individual task means 
such that learning task means ($\{\bmu_k\}_{k=1}^K$) and covariances ($\{\bSi_{\bt_k}\}_{k=1}^K$) can be decoupled via optimizing prompts $\Pb$ and the attention weight $\W$. We say the model \emph{fully decouples} mean and covariance  if the optimized attention weight $\W^\st$ is only determined by the debiased term as shown in \eqref{eqn:bias}, with prompts responsible for capturing the bias introduced by the non-zero means. 

To start with, recap from Definition~\ref{definition:multi-task} where task $k$ has probability $\pi_k$ and its task vector follows distribution $\bt_k\sim\Nc(\bmu_k,\bSi_{\bt_k})$. Following \eqref{eqn:bias}, we define the debiased and biased mixed-task covariances (variant with $\bSi_{\x}$ prior)  as follows:
\begin{subequations}\label{eqn:bt bias}
\begin{align}
\text{Debiased:}~&\scalemath{0.88}{\debias{\bSbbt}=\bSi_{\x}\sum_{k=1}^K\pi_k\E[(\bt_k-\bmu_k)(\bt_k-\bmu_k)^\top];}\\
\text{Biased:}~\quad &\scalemath{0.88}{\bias{\bStbt}
=\bSi_{\x}\sum_{k=1}^K\pi_k\E[\bt_k\bt_k^\top].}
\end{align}
\end{subequations}
Note that they satisfy  $\bSbbt=\bSi_{\x}\sum_{k=1}^K\pi_k\bSi_{\bt_k}$ and  $\bStbt=\bSi_{\x}\sum_{k=1}^K\pi_k(\bSi_{\bt_k}+\bmu_k\bmu_k^\top)$.

We first analyze the plain training setting where no additional task-specific parameters are introduced, and all $K$ tasks are mixed together. 
\begin{theorem} [Plain training]\label{thm:pretrain} Consider training a single-layer linear attention model in solving multi-task ICL problem with dataset defined in Definition~\ref{definition:multi-task} and model construction as described in Assumption~\ref{assumption:PGD}. Let the optimal solution $\W_{\pt}^\st$ (c.f.~\eqref{eqn:obj pt}) and the minimal plain training loss $\Lc_\pt^\st$ as defined in Section~\ref{sec:PGD}. 
Additionally, let $\bStbt$ 
be defined in \eqref{eqn:bt bias} and   $\Wb^\st_\pt=\bSi_{\x}\W^\st_\pt$.
Then the solution $\Wb^\star_{\pt}$ and optimal loss $\Lc^\st_\pt$ satisfy
\begin{tcolorbox}[colback=white!5!white,colframe=black!5!black,colback=white!5!white,
                  interior hidden,
                  arc=0pt,
                  boxrule=1pt,
                  boxsep=0pt,
                  left=5pt,
                  right=5pt,
                  top=5pt,]
\[
\scalemath{0.9}{\Wb^\star_{\pt}=\bias{\bStbt}\left((n+1)\bias{\bStbt}+\tr{\bias{\bStbt}}\Iden\right)^{-1},}
\]
\[
\scalemath{0.9}{\Lc^\star_{\pt} 
=\tr{\bias{\bStbt}}
-n\tr{\Wb^\star_{\pt}\bias{\bStbt}}.}
\]
\end{tcolorbox}
\end{theorem}
Note that the above solution and optimal loss are identical to those in previous work \citep{li2024fine} when considering a single-task ICL problem with with task vector following distribution $\bt\sim\Nc(0,\bStbt)$.


\begin{theorem} [Fine-tuning]\label{thm:pretrain-finetune} Suppose a pretrained model as described in Theorem~\ref{thm:pretrain} is given with $\W^\st_\pt$ being its optimal solution. Consider fine-tuning this model with task-specific prompts as defined in Definition~\ref{def:task-specific param}, and let the optimal prompt matrix $\Pb^\st_\ft$ (c.f.~\eqref{eqn:obj ft}) and the minimal fine-tuning loss $\Lc^\st_\ft$ be defined in Section~\ref{sec:PGD}. Additionally, let $\bSbbt,\bStbt$ be defined in \eqref{eqn:bt bias} and $\Wb^\st_\pt=\bSi_{\x}\W^\st_\pt$,  and define the mean matrix
\[
\M_\mu=[\bmu_1~\cdots~\bmu_K]^\top\in\R^{K\times d}.
\]
Then the solution $\Pb^\st_\ft$ and optimal loss $\Lc^\st_\ft$ satisfy 
\begin{tcolorbox}[colback=white!5!white,colframe=black!5!black,colback=white!5!white,
                  interior hidden,
                  arc=0pt,
                  boxrule=1pt,
                  boxsep=0pt,
                  left=5pt,
                  right=5pt,
                  top=3pt,]
\[
\scalemath{0.9}{\Pb^\st_\ft = \M_\mu\left((\Wb^\star_{\pt})^{-1} - n\Iden\right)\bSi_{\x},}
\]
\[
\scalemath{0.9}{\Lc^\star_\ft
=\Lc^\star_{\pt}-\tr{(\bias{\bStbt}-\debias{\bSbbt})(n\Wb^{\star}_{\pt}-\Iden)^\top(n\Wb^{\star}_{\pt}-\Iden)}.}
\]
\end{tcolorbox}
\end{theorem}
Results in Theorem~\ref{thm:pretrain-finetune} show that, fine-tuning achieves better loss than plain training, $\Lc^\st_\ft\leq\Lc^\st_\pt$, and our results provably quantize the loss difference.

\begin{theorem}[Joint training]\label{thm:joint-training} Consider training a single-layer linear attention model in solving multi-task ICL problem with dataset defined in Definition~\ref{definition:multi-task} and model construction as described in Assumption~\ref{assumption:PGD}. Let $\W_{\jt}^\st,\Pb^\st_\jt$ (c.f.~\eqref{eqn:obj jt}) be the optimal solutions and $\Lc_\jt^\st$ is the optimal joint training loss defined in Section~\ref{sec:PGD}. 
Additionally, let $\bSbbt,\bStbt,\M_\mu$ follow the same definitions as in Theorem~\ref{thm:pretrain-finetune} and define $\Wb^\st_\jt=\bSi_{\x}\W^\st_\jt$. Then the solution $(\W_{\jt}^\st,\Pb^\st_\jt)$ and optimal loss $\Lc^\st_\jt$ satisfy 
\begin{tcolorbox}[colback=white!5!white,colframe=black!5!black,colback=white!5!white,
                  interior hidden,
                  arc=0pt,
                  boxrule=1pt,
                  boxsep=0pt,
                  left=5pt,
                  right=5pt,
                  top=-3pt,]
\[
\scalemath{0.9}{\Wb^\star_{\jt} = {\debias{\bSbbt}}
\left((n+1)\debias{\bSbbt}+ \tr{\bias{\bStbt}} \Iden+\order{1}\right)^{-1},}
\]\[
\scalemath{0.9}{\Pb_\jt^\star = \M_\mu\left((\Wb^\star_{\jt})^{-1} - n\Iden\right)\bSi_{\x},}
\]
\[
\scalemath{0.9}{\Lc^\star_{\jt} =\tr{\debias{\bSbbt}}-n\tr{\Wb^\star_{\jt}{\debias{\bSbbt}}}.}
\]
\end{tcolorbox}
Here, $\order{1}$ is a  $d\times d$-sized matrix with entries being bounded by some constant value, regardless of $n$ and $d$. 
\end{theorem}

In Theorem~\ref{thm:joint-training}, for clarity, we use $\order{1}$ to represent equality up to a matrix with entries bounded by a constant. The explicit form of $\Wb^\st_\jt$ is provided in the Appendix.

 The results of Theorem~\ref{thm:pretrain-finetune} and Theorem~\ref{thm:joint-training} highlight an important commonality of the optimal prompt:
the optimal prompts $\pbb_k$ capture the mean of corresponding tasks based on the attention weight $\W^\st_\pt$ or $\W^\st_\jt$ . For a large context length $n$, $(\pbb_k)^\st_\ft$ and $(\pbb_k)^\st_\jt$ will be approximately $-n\bSi_{\x}\bmu_k$. 
Additionally,  in a finite-dimensional setting where $d<\infty$, as $n\to \infty$, the solutions $\Wb^\st_\pt$ and $\Wb^\st_\jt$ converge to $\Iden/n$ and all the optimal losses $\Lc^\st_\pt$, $\Lc^\st_\ft$ and $\Lc^\st_\jt$ approach $0$. Therefore, the benefits of prompt tuning are more apparent  for finite $n$.

\begin{corollary}
\label{corollary:ordering}
Let $\Lc^*_{\pt}$, $\Lc^*_{\ft}$, and $\Lc^*_{\jt}$ denote the optimal losses for plain training, fine-tuning, and joint training, as described in Theorems 1, 2 and 3, respectively. These losses satisfy:
\begin{align}
\Lc^*_{\jt} \leq \Lc^*_{\ft} \leq \Lc^*_{\pt}.
\end{align}
The equalities hold if and only if $\bSb_\beta = \bSp_\beta$ (c.f. (14)), which occurs when all task means $\bmu_k = \zerbb$ for $k \in [K]$. Furthermore, the loss gaps satisfy the following:
\begin{enumerate}
    \item The loss gaps scale quadratically with task mean:
    \[\scalemath{0.8}{
       \Lc^*_{\pt} - \Lc^*_{\ft} \sim \mathcal{O}\left(\frac{1}{n^2}\right)\|\Delta\|_F,\text{ } 
       \Lc^*_{\ft} - \Lc^*_{\jt} \sim \mathcal{O}\left(\frac{1}{n}\right)\|\Delta\|_F},
       \]
    where $\Delta := \bias{\bStbt}-\debias{\bSbbt} = \bSi_x\sum_{k=1}^K \pi_k\bmu_k\bmu_k^\top$.
    
    \item The ratio between gaps is: $\frac{\Lc^*_{\pt} - \Lc^*_{\ft}}{ \Lc^*_{\ft}- \Lc^*_{\jt} } \sim \mathcal{O}\left(\frac{1}{n}\right)$,
    indicating that fine-tuning provides most of the benefit in few-shot regimes (small $n$), while joint training benefits more for larger $n$.
\end{enumerate}
\end{corollary}

\subsection{Covariance-mean decoupling}\label{sec:decoupling}
For a single-layer linear attention model under Assumption~\ref{assumption:PGD}, where $\W \in \R^{d \times d}$ captures all the statistics, the plain training loss $\Lc^\star_\pt$ is determined by the second-order moment of the task parameters $\bt_k$:
$$
\E[\bt_k \bt_k^\top] = \bSi_{\bt_k} + \bmu_k \bmu_k^\top,
$$
which represents the biased covariance of task $k$. Consequently, $\Lc^\star_\pt$ can be viewed as a function of this biased variable, defined as $\bias{\bStbt}$ (see Theorem~\ref{thm:pretrain}), which affects both the terms in $\W^\st_\pt$ and those that appear directly in $\Lc^\star_\pt$ but outside of $\W^\st_\pt$.

When all tasks have zero mean, i.e., $\E[\bt_k] = \bmu_k = \mtx{0}$ for all $k \in [K]$, leading to $\bSbmu = \zerbb_{d \times d}$, the optimal losses for pretraining, fine-tuning, and joint training become identical:
\begin{align}
    \bmu_k = \mtx{0}, \, k \in [K] \Rightarrow \Lc^\star_\pt = \Lc^\star_\ft = \Lc^\star_\jt.
    \label{eqn:zero_mean_loss}
\end{align}
In this case, the fine-tuned prompts remain as zero vectors, learning nothing during fine-tuning (as shown in Theorem~\ref{thm:pretrain-finetune} and Theorem~\ref{thm:joint-training}). This implies that any differences in loss arise from the ability of the trainable prompts to \textbf{decouple} (i.e., remove task-mean related bias terms) from $\bias{\bStbt}$ to obtain $\debias{\bSbbt}$. Specifically, when all task means are zero, this decoupling is nullified, resulting in no difference in losses across the training methods.

When the task means $\E[\bt_k] = \bmu_k$ are non-zero, the decoupling effect varies between different training settings. In joint training, the simultaneous optimization of prompts and attention weights enables decoupling in both $\Lc^\st_\jt$ and $\W^\st_\jt$, while in fine-tuning, only the biased terms in $\W^\st_\ft$ are decoupled. As a result, joint training achieves greater decoupling than fine-tuning. According to Corollary~\ref{corollary:ordering}, when a loss is more directly influenced by the biased covariance $\bias{\bStbt}$, it tends to be higher, whereas reducing the influence of $\bias{\bStbt}$ through decoupling generally leads to a lower loss.

These findings suggest that introducing additional task-specific trainable parameters into the single-layer linear attention model, combined with joint optimization, can effectively reduce the bias in mixed-task covariance, thereby improving performance.

%% file: sec/vector_head.tex
\section{Fully-decoupled Loss}
\label{sec:fully-decoupled}
In Section~\ref{sec:main}, we focus on cases where model parameters are constructed according to Assumption \ref{assumption:PGD} and analyze the loss landscapes for plain training, finetuning, and joint training. However, our results indicate that with a shared head $\ab:=\WV\hb$, none of these methods fully decouple the mean and covariance. To address this, we introduce an alternative approach that allows each task to have its own specific linear prediction head $\hb_k\in\R^{d+1}$. It is worth noting that using separate heads for different tasks is a common practice in the general multi-task learning literature \citep{caruana1997multitask,zhang2021survey,li2023provable}. Our results demonstrate that optimizing task-specific prompts, heads, and attention weights leads to a fully decoupled loss (Theorem~\ref{thm:fully_decoupled}).
\begin{definition}[Task-specific heads]
\label{def:task-specific head} 
Given $K$ tasks, let $\{\hb_k\}_{k=1}^K\subset\R^{d+1}$ represent their corresponding trainable linear prediction heads. Recalling the input sequence and prediction from \eqref{def Z k} and \eqref{eqn:LinearAttn}, the prediction for task $k$ returns
\begin{align}
\scalemath{0.9}{\tilde f_{\att}(\Zk)=(\z^\top\WQ\WK^\top(\Zk)^\top)\M\Zk\WV\hb_k.}\label{eqn:LinearAttn_task}
\end{align}
\end{definition}
%
%
%
Recap ICL problem from Definition~\ref{definition:multi-task}, $\Zk$ from Definition~\ref{def:task-specific param} and loss function from \eqref{obj}, the ICL objective considering task-specific prompts and heads is:
\begin{align}
&\scalemath{0.9}{\tilde\Lc_\att^\st=\min_{(\pb_k,\hb_k)_{k=1}^K,\W_{k,q,v}}\Lc(\tilde f_\att)}\label{eqn:obj att h}\\
    &\scalemath{0.9}{\textnormal{where }\Lc(\tilde f_\att)=\sum_{k=1}^K\pi_k\E_{\Z,y\sim\Dc_k}[(\tilde f_\att(\Zk)-y)^2].}\nn
\end{align}
Here, the search space for $\pb_k,\hb_k$ is $\R^{d+1}$ and  the search space for $\W_{k,q,v}$ is $\R^{(d+1)\times(d+1)}$.

Next, given task $k$ with mean $\bmu_k$, we introduce debiased preconditioned gradient descent (PGD) predictor as follows: 
\begin{align*}
    \scalemath{0.9}{\tilde f_\pgd(\Zk)=\x^\top\W\X^\top(\y-\X\bmu_k)+\x^\top\bmu_k.}
\end{align*}
Note that for any task $k$, we have $\E[y_i-\x_i^\top\bmu_k]=0$, and therefore, we expect $\tilde f_\pgd(\Zk)-\x^\top\bmu_k$ to predict unbiased label $y-\x^\top\bmu_k$.
The corresponding PGD objective is defined as:
\begin{align}
&\scalemath{0.9}{\tilde\Lc_\pgd^\st=\min_{\W\in\R^{d\times d}}\Lc(\tilde f_\pgd)\label{eqn:obj pgd h}}\\
    &\scalemath{0.9}{\textnormal{where }\Lc(\tilde f_\pgd):=\sum_{k=1}^K\pi_k\E_{\Z,y\sim\Dc_k}[(\tilde f_\pgd(\Zk)-y)^2].}\nn
\end{align}
The following proposition establishes the equivalence between optimizing single layer linear attention (c.f.~\eqref{eqn:obj att h}) and one step of PGD predictor (c.f.~\eqref{eqn:obj pgd h}).
\begin{proposition}\label{prop h}
Consider the multi-task ICL data as described in Definition~\ref{definition:multi-task} and let $\tilde\Lc_\att^\st$ and $\tilde\Lc_\pgd^\st$ be the optimal linear attention and debiased preconditioned gradient descent losses as presented in \eqref{eqn:obj att h} and \eqref{eqn:obj pgd h}, respectively. Then, $\tilde\Lc_\att^\st=\tilde\Lc_\pgd^\st$.
\end{proposition}
Considering the single-task and zero-mean setting, Proposition~\ref{prop h} aligns with the findings of previous work \citep{ahn2024transformers,li2024fine}. Our results further confirm the necessity of both task-specific prompts and heads in effectively decoupling the influence of non-zero means from the data.

Then, based on the Proposition~\ref{prop h}, we are able to analyze the optimization landscape of \eqref{eqn:obj att h} via studying \eqref{eqn:obj pgd h}.
\begin{theorem}\label{thm:fully_decoupled}
 Consider the multi-task ICL problem with dataset defined in Definition~\ref{definition:multi-task}. Let $\W_\pgd^\st:=\arg\min_{\W}\Lc(\tilde f_\pgd)$ following \eqref{eqn:obj pgd h}. Define $\bSbbt$ in \eqref{eqn:bt bias} and let $\Wb_\pgd^\st=\bSi_{\x}\W_\pgd^\st$. Then the solution $\Wb_\pgd^\st$ and optimal loss $\tilde\Lc^\st_\pgd$ (c.f.~\eqref{eqn:obj pgd h}) satisfy 
\begin{tcolorbox}[colback=white!5!white,colframe=black!5!black,colback=white!5!white,
                  interior hidden,
                  arc=0pt,
                  boxrule=1pt,
                  boxsep=0pt,
                  left=5pt,
                  right=5pt,
                  top=5pt,]
\[
\scalemath{0.9}{\Wb_\pgd^\star=\debias{\bSbbt}\left((n+1)\debias{\bSbbt}+\tr{\debias{\bSbbt}}\Iden\right)^{-1},}
\]
\[
\scalemath{0.9}{\tilde\Lc^\star_{\pgd} 
=\tr{\debias{\bSbbt}}
-n\tr{\Wb_\pgd^\star\debias{\bSbbt}}.}
\]
\end{tcolorbox}
\end{theorem}
See Appendix for a proof. 

Comparing Theorem~\ref{thm:fully_decoupled} with Theorem~\ref{thm:pretrain}, it is evident that $\tilde\Lc^\star_{\pgd}$ represents a fully decoupled loss, demonstrating the clear benefit of adding task-specific heads. While Theorem~\ref{thm:pretrain} provides an upper bound $\Lc^\st_\pt$ for the multi-task ICL loss, the proof of Theorem~\ref{thm:fully_decoupled} establishes that $\tilde\Lc^\star_{\pgd}$ serves as the lower bound for a multi-task ICL loss in a single-layer linear attention model.

%% file: sec/exp.tex
\input{AISTATS_figs/exp_reduced_model}
\section{Experiments}
We conduct experiments on synthetic datasets to validate our theoretical assumptions and explore the behavior of single-layer linear attention models with various trainable parameters under different training settings.

\textbf{Experimental Setting.} We train single-layer attention models to solve $K$-task, $d$-dimensional linear regression ICL in a noise-free meta-learning setup for consistency with the main paper’s theorems, deferring noisy results to the Appendix. For each context length $n$, an independent model is trained for $20,000$ iterations with a batch size of $8192$ using the Adam optimizer (learning rate $10^{-3}$). 

To ensure robustness, each training process is repeated $50$ times with independent initializations, and the minimal test risk among these trials is reported. Theoretical predictions in the plots are based on the theorems in Section~\ref{sec:main}, and all results are normalized by $\E[\|y\|^2]$.

\textbf{Validation of the preconditioning.} In order to support Assumption~\ref{assumption:PGD} in Section~\ref{sec:PGD}, we train an unconstrained $\WK, \WQ, \WV$-parameterized model (with a shared head $\hb$) which is of an alternative form~\ref{eqn:LinearAttn}, as well as a reduced model~\ref{eqn:pgd} derived from Assumption~\ref{assumption:PGD}, in order to check the alignment of their loss landscape. 
We configure the experiment with the following choice:
\[
\begin{aligned}
    &d=10, K=2, \M_\mu = \begin{bmatrix} 
    1.7 \cdot \onebb_{10} & 
    -1.3 \cdot \onebb_{10} 
    \end{bmatrix}, \\
    &\bSi_{\bt_1} = \bSi_{\bt_2} = \Iden_{10}, \quad \pi_1 = 0.3, \quad\pi_2 = 0.7.
\end{aligned}
\]
\[
\begin{aligned}
    &d=10, K=2, \M_\mu = \begin{bmatrix} 
    \zerbb_{10} & 
    \zerbb_{10} 
    \end{bmatrix}, \\
    &\bSi_{\bt_1} = \Iden_{10}, \bSi_{\bt_2} = 2\cdot\Iden_{10}, \quad \pi_1 =\pi_2 = 0.5.
\end{aligned}
\]
As shown in Figure~\ref{fig:reduced_1},~\ref{fig:reduced_2}, the alignment of the dashed and solid lines across the plain training, fine-tuning, and joint training settings suggests that Assumption~\ref{assumption:PGD} results in a simple yet effective reduced model. The performance of this reduced model aligns closely with that of optimizing a single-layer linear attention model without the constraint imposed by this assumption, indicating its reasonableness. We also validated that the performance of different training settings of the unconstrained  model and the reduced model will be aligned given a zero task mean scenario, which is stated in Section~\ref{sec:decoupling}.

Note that the joint optimization of the task-specific head, prompt, and model attention weights is not included in this experiment, although it could potentially support Theorem~\ref{thm:fully_decoupled}. When the task-specific head is included in the joint training, Assumption~\ref{assumption:PGD} is no longer needed to derive the reduced form, making validation of Assumption~\ref{assumption:PGD} under such settings unnecessary.

\textbf{Validation of the theorems.} Given the previous experimental results supporting Assumption~\ref{assumption:PGD}, which in turn validate Theorems~\ref{thm:pretrain}, \ref{thm:pretrain-finetune}, and \ref{thm:joint-training}, these theorems are primarily supported by empirical evidence. Similarly, we train an unconstrained $\WK, \WQ, \WV$-parameterized model (with a shared head $\hb$) to assess the alignment between its loss landscape and our theoretical predictions from Theorems~\ref{thm:pretrain}, \ref{thm:pretrain-finetune}, and \ref{thm:joint-training}. The experiments are configured with the following settings:
\[
\begin{aligned}
    &d=10, K=2, \M_\mu = \begin{bmatrix} 
    2.1\cdot\onebb_{10} & 
    -0.9\cdot\onebb_{10} 
    \end{bmatrix}, \\
    &\bSi_{\bt_1} = \bSi_{\bt_2} = \Iden_{10}, \quad \pi_1 =0.3, \pi_2 = 0.7.
\end{aligned}
\]

    As shown in Figure~\ref{fig:theoretical}, the alignment of the dashed and solid lines across the plain training, fine-tuning, and joint training settings suggests that our theoretical result can predict the multi-task linear regression ICL loss accurately.

\textbf{Benefits of Additional Task-Specific Parameters.} We assess the impact of adding task-specific heads to a single-layer linear attention model with task-specific prompts. Specifically, we compare the loss $\tilde\Lc_\att^\st$ from jointly training $\W, \Pb, \Hb$ to the loss $\Lc_\jt^\st$ from jointly training $\W, \Pb$, using $\Lc_\pt^\st$ as a baseline, where only $\W$ is optimized. Theoretical curves are shown only for Theorem~\ref{thm:fully_decoupled}, as the others have been validated in Theorems~\ref{thm:pretrain}, \ref{thm:pretrain-finetune}, and \ref{thm:joint-training}. The experiments are configured as follows:
\[
\begin{aligned}
    &d=10, K=2, \M_\mu = \begin{bmatrix} 
    1.4\cdot\onebb_{10} & 
    -0.6\cdot\onebb_{10} 
    \end{bmatrix}, \\
    &\bSi_{\bt_1} =\Iden_{10}, \bSi_{\bt_2} =2\cdot \Iden_{10}, \quad \pi_1 =0.3, \pi_2 = 0.7.
\end{aligned}
\]
There is a clear, gradual improvement from adding more task-specific parameters, as shown in Figure~\ref{fig:tunable_head}.

\textbf{Discussions.} We analyze the impact of task-specific parameters in multi-task ICL settings. Our work provides theoretical guarantees for joint training and pretrain $\rightarrow$ finetune approaches. We introduce a covariance-mean decoupling mechanism for optimal ICL: Task-specific parameters learn the task mean prediction and attention weights learn the variance. Experimental results support our theoretical analysis.

%% file: AISTATS_figs/exp_reduced_model.tex
\begin{figure*}[t]
\vspace{-10pt}
\centering
~\hspace{-15pt}
\subfloat[]{\begin{tikzpicture}
\node at (-3.2,-.4)[anchor=west]{\includegraphics[width=0.25\textwidth]{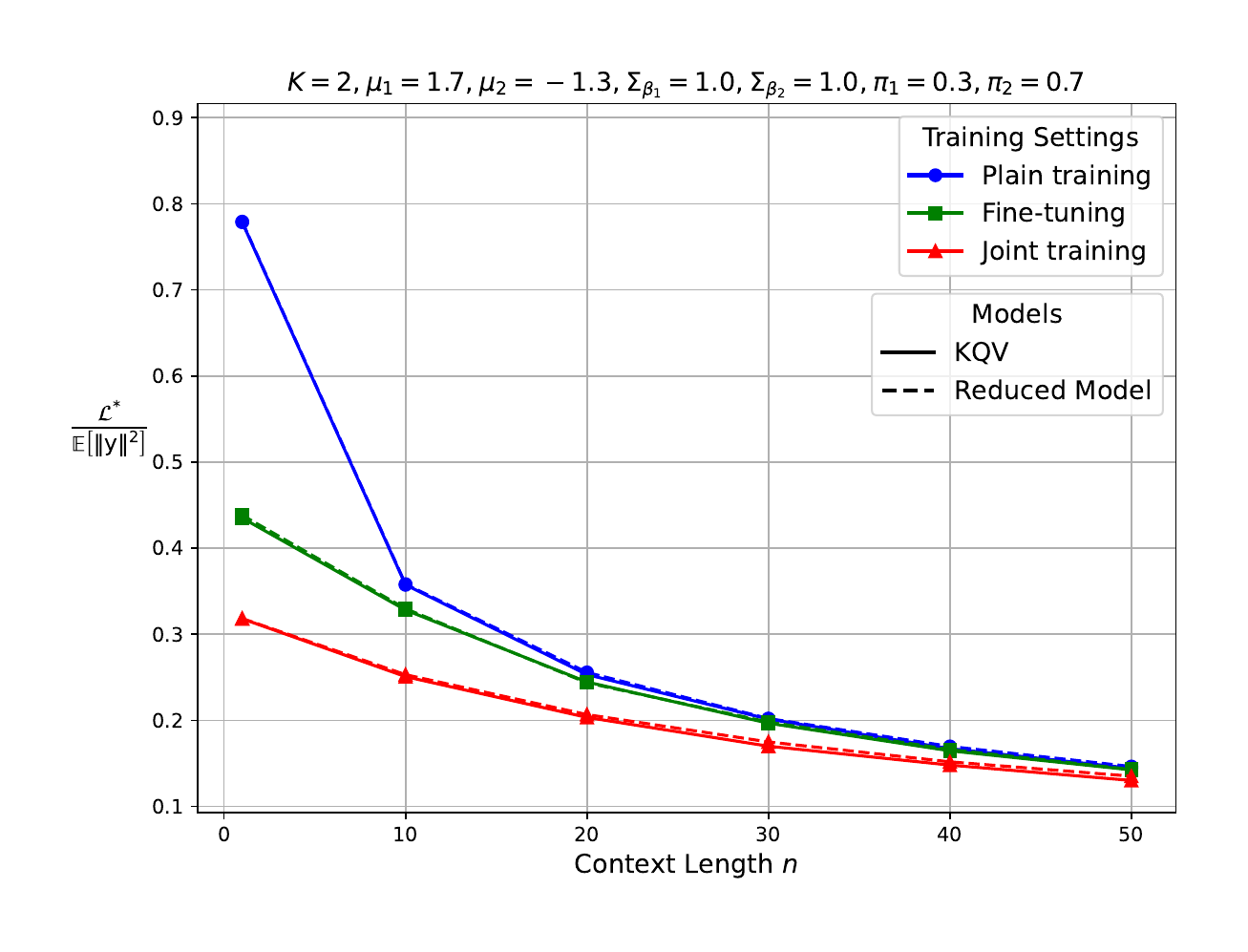}};
\node at (-1.1,1.35)[anchor=west][scale=0.6,opacity=1]{};
\node at (-2.9,-1.6)[anchor=west][scale=0.6,rotate=90,opacity=1]{ };
\label{fig:reduced_1}
\end{tikzpicture}
}
~\hspace{-15pt}
\subfloat[]{\begin{tikzpicture}
\node at (-3.2,-.4)[anchor=west] {\includegraphics[width=0.25\textwidth]{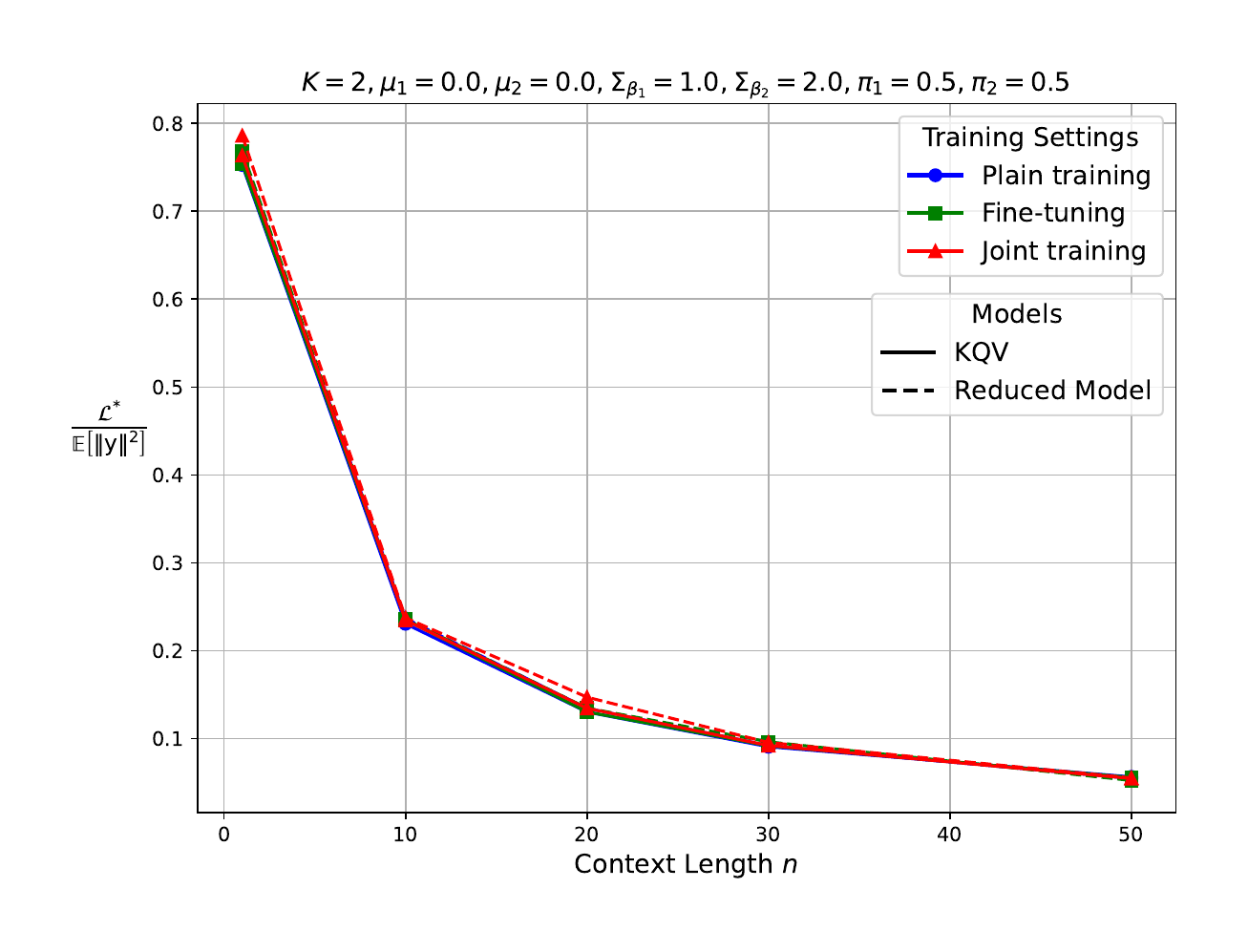}};
\node at (-0.7,1.35)[anchor=west][scale=0.6,opacity=1]{};
\label{fig:reduced_2}
\end{tikzpicture}
}
~\hspace{-15pt}
\subfloat[]{\begin{tikzpicture}
\node at (-3.2,-.4)[anchor=west]{\includegraphics[width=0.25\textwidth]{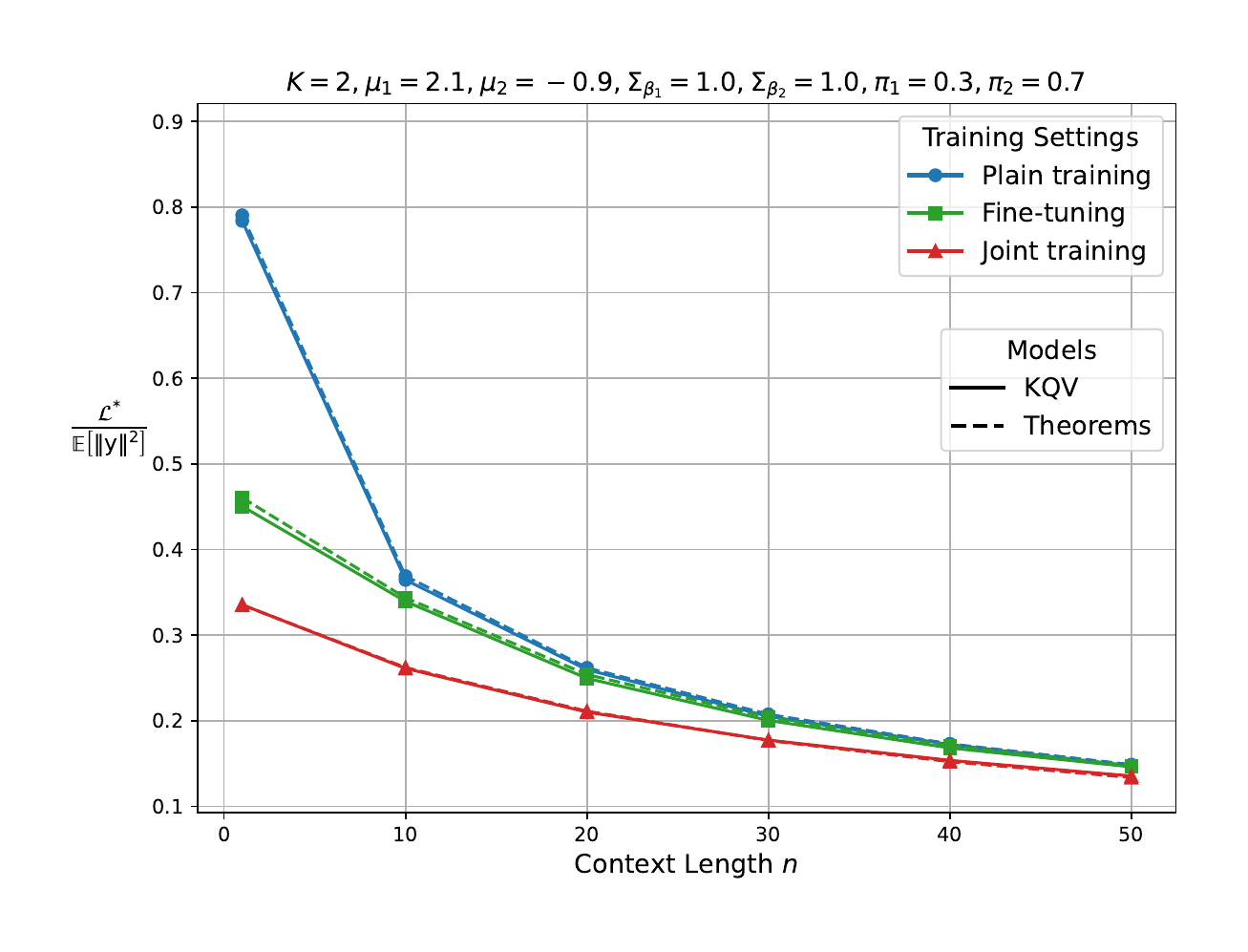}};
\node at (-1.1,1.35)[anchor=west][scale=0.6,opacity=1]{};
\node at (-2.9,-1.6)[anchor=west][scale=0.6,rotate=90,opacity=1]{ };
\label{fig:theoretical}
\end{tikzpicture}
}
~\hspace{-15pt}
\subfloat[]{\begin{tikzpicture}
\node at (-3.2,-.4)[anchor=west] {\includegraphics[width=0.25\textwidth]{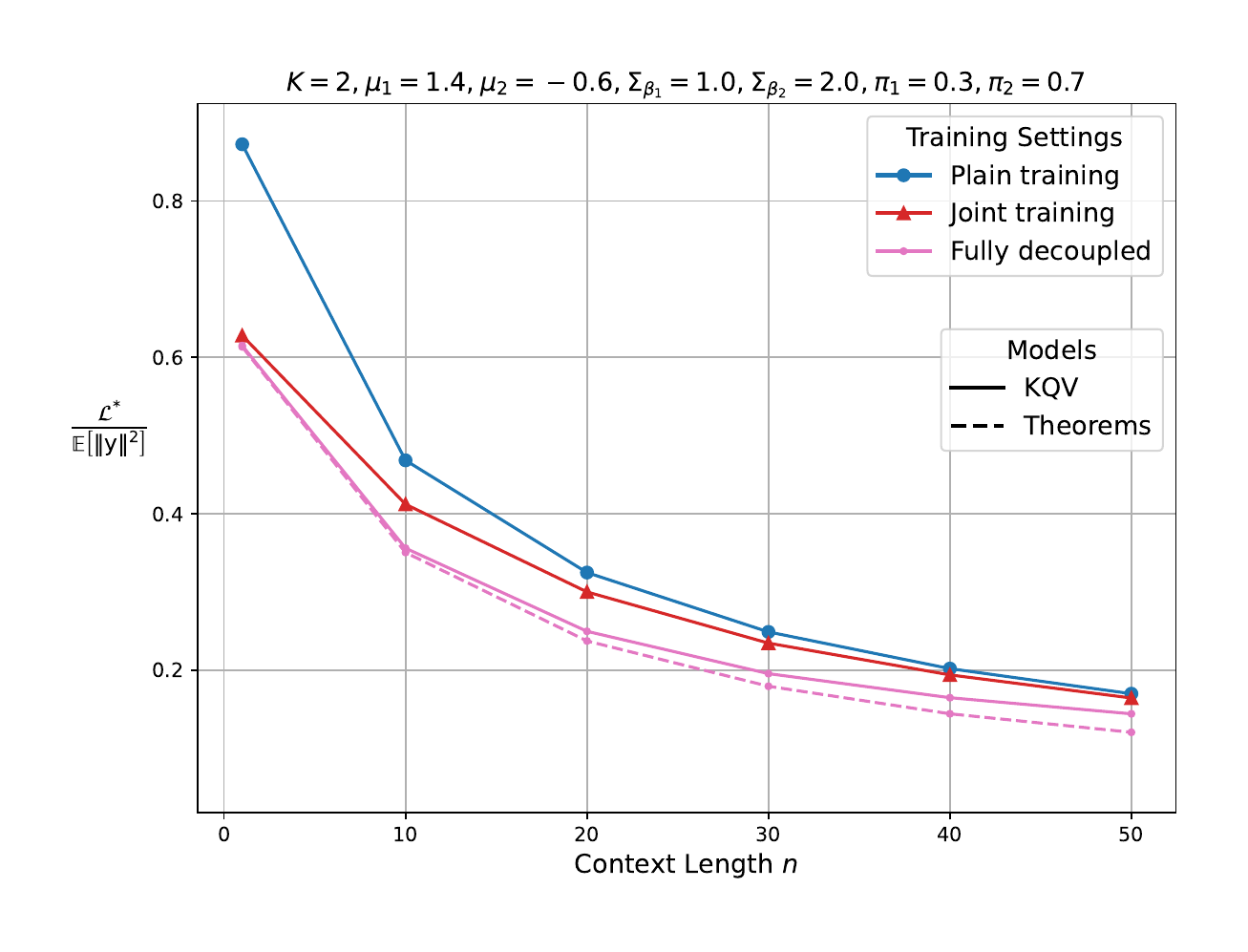}};
\node at (-0.7,1.35)[anchor=west][scale=0.6,opacity=1]{};
\label{fig:tunable_head}
\end{tikzpicture}
}
\caption[caption]{Experimental results across various settings: (a) Performance of unconstrained $\WK,\WQ,\WV$-parameterized linear attention model and reduced model, with non-zero task mean. (b) Performance of unconstrained $\WK,\WQ,\WV$-parameterized linear attention model and reduced model, with zero task mean. (c) Performance of unconstrained $\WK,\WQ,\WV$-parameterized linear attention model and theoretical prediction, with non-zero task mean. (d)  Performance of unconstrained $\WK,\WQ,\WV$-parameterized linear attention model, with different numbers of task-specific trainable parameters.} 
\vspace{-0.17in}
\end{figure*}

%% file: sec/acknowledgement.tex
\section*{Acknowledgments}
This work was partially supported under NSF grants CCF2046816, CCF-2403075, CCF-2008020, and the Office of Naval Research grants N000142412289 and N000141812252. Additionally, research was sponsored by the OUSD (R\&E)/RT\&L and was accomplished under Cooperative Agreement Number W911NF-20-2-0267. The views and conclusions contained in this document are those of the authors and should not be interpreted as representing the official policies, either expressed or implied, of the ONR, ARL and OUSD(R\&E)/RT\&L or the U.S. Government. The U.S. Government is authorized to reproduce and distribute reprints for Government purposes notwithstanding any copyright notation herein. This work was also partially supported by gifts from Open Philanthropy, Amazon Research, and Google Research.

%% file: app/checklist.tex
\section*{Checklist}



 \begin{enumerate}

 \item For all models and algorithms presented, check if you include:
 \begin{enumerate}
   \item A clear description of the mathematical setting, assumptions, algorithm, and/or model. [Yes]
   \item An analysis of the properties and complexity (time, space, sample size) of any algorithm. [Not Applicable]
   \item (Optional) Anonymized source code, with specification of all dependencies, including external libraries. [Yes]
 \end{enumerate}

 \item For any theoretical claim, check if you include:
 \begin{enumerate}
   \item Statements of the full set of assumptions of all theoretical results. [Yes]
   \item Complete proofs of all theoretical results. [Yes]
   \item Clear explanations of any assumptions. [Yes]     
 \end{enumerate}

 \item For all figures and tables that present empirical results, check if you include:
 \begin{enumerate}
   \item The code, data, and instructions needed to reproduce the main experimental results (either in the supplemental material or as a URL). [Yes]
   \item All the training details (e.g., data splits, hyperparameters, how they were chosen). [Yes]
         \item A clear definition of the specific measure or statistics and error bars (e.g., with respect to the random seed after running experiments multiple times). [Yes]
         \item A description of the computing infrastructure used. (e.g., type of GPUs, internal cluster, or cloud provider). [Not Applicable]
 \end{enumerate}

 \item If you are using existing assets (e.g., code, data, models) or curating/releasing new assets, check if you include:
 \begin{enumerate}
   \item Citations of the creator If your work uses existing assets. [Not Applicable]
   \item The license information of the assets, if applicable. [Not Applicable]
   \item New assets either in the supplemental material or as a URL, if applicable. [Not Applicable]
   \item Information about consent from data providers/curators. [Not Applicable]
   \item Discussion of sensible content if applicable, e.g., personally identifiable information or offensive content. [Not Applicable]
 \end{enumerate}

 \item If you used crowdsourcing or conducted research with human subjects, check if you include:
 \begin{enumerate}
   \item The full text of instructions given to participants and screenshots. [Not Applicable]
   \item Descriptions of potential participant risks, with links to Institutional Review Board (IRB) approvals if applicable. [Not Applicable]
   \item The estimated hourly wage paid to participants and the total amount spent on participant compensation. [Not Applicable]
 \end{enumerate}

 \end{enumerate}

%% file: AISTATS_supp.tex
\setcounter{lemma}{0}
\setcounter{theorem}{0}
\setcounter{definition}{0}
\renewcommand{\theequation}{A\arabic{equation}}
\setcounter{figure}{0}
\renewcommand{\thefigure}{A\arabic{figure}}
\setcounter{equation}{0}
\setcounter{page}{0}
\setcounter{corollary}{0}
\setcounter{proposition}{0}
\DoToC\vfill
\section{Lemmas}
\input{app/Appendix/lemma}

\section{Proofs for Section~\ref{sec:main}}
\input{app/Appendix/theorem}
\section{Proofs for Section~\ref{sec:fully-decoupled}}

\input{app/Appendix/proof_fully_decoupled}
\section{Additional experiments: noisy label and non-isotropic covariance}
\input{app/Appendix/noisy}

\section{Additional experiments: multi-layer linear attention models}
\input{app/Appendix/multilayer}

%% file: app/Appendix/lemma.tex
\begin{lemma} Suppose $\X$ is a $n \times d$ matrix, each column of which is independently drawn from a $d$-variate Gaussian distribution with zero mean:
$$\X=[\x_1~\dots~\x_n]^\top,~\text{where}~\x_i\sim\Nc(0,\bSi_{\x})\in\R^d,i\in[n].$$
For a constant matrix $\A\in\R^{d\times d}$, the following expectation can be determined by:
$$\E\left[\X^\top\X\A\X^\top\X\right]=n\tr{\bSi_{\x}\A}\bSi_{\x}+n(n+1)\bSi_{\x}\A\bSi_{\x}.$$
\label{lemma:wishart}
\end{lemma}
\begin{proof}
We begin by expressing $\X^\top \X$ as a sum over its columns:
\[
\X^\top \X = \sum_{i=1}^n \x_i \x_i^\top.
\]
Therefore,
\[
\X^\top \X \A \X^\top \X = \left( \sum_{i=1}^n \x_i \x_i^\top \right) \A \left( \sum_{j=1}^n \x_j \x_j^\top \right) = \sum_{i=1}^n \sum_{j=1}^n \x_i \x_i^\top \A \x_j \x_j^\top.
\]
Taking expectations on both sides, we have:
\[
\E[\X^\top \X \A \X^\top \X] = \sum_{i=1}^n \sum_{j=1}^n \E[\x_i \x_i^\top \A \x_j \x_j^\top].
\]
Since the vectors $\x_i$ are independent and identically distributed, we can split the sum into terms where $i = j$ and $i \ne j$:
\[
\E[\X^\top \X \A \X^\top \X] = \sum_{i=1}^n \E[\x_i \x_i^\top \A \x_i \x_i^\top] + \sum_{i \ne j} \E[\x_i \x_i^\top \A \x_j \x_j^\top].
\]
For $i \ne j$, independence implies:
\[
\E[\x_i \x_i^\top \A \x_j \x_j^\top] = \E[\x_i \x_i^\top] \A \E[\x_j \x_j^\top] = \bSi_{\x} \A \bSi_{\x}.
\]
There are $n(n - 1)$ such terms. For $i = j$, we compute:
\[
\E[\x_i \x_i^\top \A \x_i \x_i^\top] = \E[(\x_i^\top \A \x_i) \x_i \x_i^\top].
\]
Using Isserlis' theorem for zero-mean Gaussian vectors, we have:
\[
\E[(\x^\top \A \x) \x \x^\top] = \tr{\A \bSi_{\x}}\, \bSi_{\x} + 2 \bSi_{\x} \A \bSi_{\x}.
\]
Thus, summing over $n$ terms:
\[
\sum_{i=1}^n \E[\x_i \x_i^\top \A \x_i \x_i^\top] = n \left( \tr{\A \bSi_{\x}}\, \bSi_{\x} + 2 \bSi_{\x} \A \bSi_{\x} \right).
\]
Adding all terms together:
\begin{align*}
\E[\X^\top \X \A \X^\top \X] &= n \left( \tr{\A \bSi_{\x}}\, \bSi_{\x} + 2 \bSi_{\x} \A \bSi_{\x} \right) + n(n - 1) \bSi_{\x} \A \bSi_{\x} \\
&= n \tr{\A \bSi_{\x}}\, \bSi_{\x} + \left( 2n + n(n - 1) \right) \bSi_{\x} \A \bSi_{\x} \\
&= n \tr{\A \bSi_{\x}}\, \bSi_{\x} + n(n + 1) \bSi_{\x} \A \bSi_{\x}.
\end{align*}
This completes the proof.
\end{proof}

\begin{lemma} Suppose $\X$ is a $n \times d$ matrix, each column of which is independently drawn from a $d$-variate Gaussian distribution with zero mean:
$$\X=[\x_1~\dots~\x_n]^\top,~\text{where}~\x_i\sim\Nc(0,\bSi_{\x})\in\R^d,i\in[n].$$
For a zero-mean Gaussian variable sampled independently $\mtx{\xi}\sim\Nc(0,\sigma^2\Iden_n)\in\R^n$, the following expectation moment can be determined by:
$$\E\left[\X^\top\mtx{\xi}\mtx{\xi}^\top\X\right]=n\sigma^2\bSi_{\x}.$$
\label{lemma:noise}
\end{lemma}
\begin{proof} Using the independence of $\X$ and the entries of $\mtx{\xi}=[\xi_1~\cdots~\xi_n]^\top$, $\E[\xi_i\xi_j]=\sigma^2\delta_{ij}$ (where $\delta_{ij}$ is the Kronecker delta, equal to $1$ if $i = j$ and $0$ otherwise):
\[
\begin{aligned}
\E\left[\X^\top\mtx{\xi}\mtx{\xi}^\top\X\right]&=\E\left[\sum_{i=1}^n\sum_{j=1}^n\xi_i\xi_j\x_i\x_j^\top\right]\\
&=\sum_{i=1}^n\E[\xi_i\xi_i]\E[\x_i\x_i^\top]\\
&=n\sigma^2\bSi_{\x}.
\end{aligned}
\]
\end{proof}

\begin{lemma}\label{lemma:derivative_2nd}
Let $\W \in \mathbb{R}^{d \times d}$, and $\A, \B \in \mathbb{R}^{d \times d}$ be constant matrices. Then,
\[
\frac{\partial}{\partial \W} \tr{\W \A \W^\top \B} = \B^\top \W \A^\top + \B \W \A.
\]
\end{lemma}
\begin{proof}
We will compute the derivative $\frac{\partial}{\partial \W} \tr{\W \A \W^\top \B}$ using an element-wise approach.

First, expand the trace function:
\[
\tr{\W \A \W^\top \B} = \sum_{i=1}^d (\W \A \W^\top \B)_{ii}.
\]

Using the definition of matrix multiplication, we have:
\[
(\W \A)_{ij} = \sum_{k=1}^d W_{ik} A_{kj},
\]
\[
(\W^\top \B)_{ji} = \sum_{l=1}^d W_{lj} B_{li}.
\]

Therefore,
\[
(\W \A \W^\top \B)_{ii} = \sum_{j=1}^d (\W \A)_{ij} (\W^\top \B)_{ji} = \sum_{j=1}^d \left( \sum_{k=1}^d W_{ik} A_{kj} \right) \left( \sum_{l=1}^d W_{lj} B_{li} \right).
\]

Thus, the trace becomes:
\[
\tr{\W \A \W^\top \B} = \sum_{i=1}^d \sum_{j=1}^d \sum_{k=1}^d \sum_{l=1}^d W_{ik} A_{kj} W_{lj} B_{li}.
\]

We need to compute the derivative with respect to $W_{pq}$:
\[
\frac{\partial}{\partial W_{pq}} \tr{\W \A \W^\top \B}= \sum_{i=1}^d \sum_{j=1}^d \sum_{k=1}^d \sum_{l=1}^d \frac{\partial}{\partial W_{pq}} \left( W_{ik} A_{kj} W_{lj} B_{li} \right).
\]

Note that $A_{kj}$ and $B_{li}$ are constants.

We have:
\[
\frac{\partial}{\partial W_{pq}} W_{ik} = \delta_{ip} \delta_{kq},
\]
\[
\frac{\partial}{\partial W_{pq}} W_{lj} = \delta_{lp} \delta_{jq},
\]
where $\delta_{ij}$ is the Kronecker delta, equal to $1$ if $i = j$ and $0$ otherwise.

Therefore,
\[
\frac{\partial}{\partial W_{pq}} \left( W_{ik} A_{kj} W_{lj} B_{li} \right) = \left( \delta_{ip} \delta_{kq}  W_{lj} + W_{ik}  \delta_{lp} \delta_{jq} \right) A_{kj}B_{li}.
\]

Then:
\begin{align*}
\frac{\partial}{\partial W_{pq}} \tr{\W \A \W^\top \B}&= \sum_{i=1}^d \sum_{j=1}^d \sum_{k=1}^d \sum_{l=1}^d \frac{\partial}{\partial W_{pq}} \left( W_{ik} A_{kj} W_{lj} B_{li} \right)\\
&=\sum_{i=1}^d \sum_{j=1}^d \sum_{k=1}^d \sum_{l=1}^d \left( \delta_{ip} \delta_{kq}  W_{lj} + W_{ik}  \delta_{lp} \delta_{jq} \right) A_{kj}B_{li}\\
&= \sum_{j=1}^d  \sum_{l=1}^dW_{lj}A_{qj}B_{lp}+\sum_{i=1}^d \sum_{k=1}^dW_{ik}A_{kq}B_{pi}\\
&=(\A\W^\top\B)_{qp}+(\B\W\A)_{pq}\\
&=(\B^\top\W\A^\top+\B\W\A)_{pq}
\end{align*}

Since this holds for all elements $(p, q)$, in matrix form, we have:
\[
\frac{\partial}{\partial \W} \tr{\W \A \W^\top \B} = \B^\top \W \A^\top + \B \W \A.
\]
\end{proof}
\begin{lemma}[Reduced form] Denote the output sequence of the attention layer as $\att(\Z)$. Then under Assumption~\ref{assumption:PGD}, the output becomes 

\end{lemma}
\begin{proof}
    \[
    \begin{aligned}
    \hat y&=\att(\Z)_{(n+1,d+1)}=\mtx{e}_{n+1}^\top\att(\Z)\mtx{e}_{d+1}\\
    &=\mtx{e}_{n+1}^\top(\Z\WQ\WK^\top(\Z)^\top)\M\Z\WV\mtx{e}_{d+1}\\
    &=(\mtx{e}_{n+1}^\top\Z)\underbrace{(\WQ\WK^\top)}_{=\A}(\Z^\top\M\Z)\underbrace{\WV\mtx{e}_{d+1}}_{=\ab}\\
    &=(\mtx{e}_{n+1}^\top\Z)\A(\Z^\top\M\Z)\ab\\
    &=\begin{bmatrix}
    \x\\0
    \end{bmatrix}^\top\A\begin{bmatrix}
    \X^\top\X~\X^\top\y\\
    \y^\top\X~\y^\top\y
    \end{bmatrix}\ab\\
    \end{aligned}
    \]
Under Assumption~\ref{assumption:PGD}
    \[
        \A = \begin{bmatrix} 
        \W_{d\times d} & \zerbb_{d \times 1} \\
        *_{1\times d} & *_{1\times 1} 
        \end{bmatrix},~\\
        \ab = \begin{bmatrix} 
         \zerbb_{d\times 1} \\
         1_{1\times 1} 
        \end{bmatrix},
    \]
The output finally reduced to
    \[
\begin{aligned}
        \hat y&= \begin{bmatrix}
    \x\\0
    \end{bmatrix}^\top\begin{bmatrix} 
        \W & \zerbb \\
        * & * 
        \end{bmatrix}\begin{bmatrix}
    \X^\top\X~\X^\top\y\\
    \y^\top\X~\y^\top\y
    \end{bmatrix}\begin{bmatrix} 
         \zerbb \\
         1 
        \end{bmatrix}\\
    &=\x^\top\W\X^\top\y.
\end{aligned}
    \]
\end{proof}

%% file: app/Appendix/theorem.tex
We consider an in-context learning (ICL) problem with demonstrations $(\x_i,y_i)_{i=1}^{n+1}$, and the input sequence $\Z$ is defined by removing $y_{n+1}$ as follows:
\begin{align}\label{def Z app}
\Z&=[\z_1~\dots~\z_n~\z]^\top=\begin{bmatrix}
    \x_1&\dots&\x_n&\x\\
    y_1&\dots&y_n&0
\end{bmatrix}^\top=\begin{bmatrix}
    \X^\top&\x\\
    \y^\top&0
\end{bmatrix}^\top\in\R^{(n+1)\times(d+1)}.
\end{align}
 
 Here, $\z=[\x^\top~0]^\top$ is the query token where $\x:=\x_{n+1}$, and $\X=[\x_1~\cdots~\x_n]^\top\in\R^{n\times d}$, $\y=[y_1~\cdots~y_n]^\top\in\R^n$. 
 Then, we aim for a sequence model to predict the associated label  $y:=y_{n+1}$ of the given input sequence $\Z$. In this work, we consider the following data generation of $(\Z,y)$. We will refer to $(\X, \y)$, $\x$, and $y$ as contexts, query feature and the label to predict, respectively.
\begin{definition}[Single-task ICL]\label{def:app_singletask}
 Given a task mean $\bmu\in\R^d$, and covariances $\bSi_{\x},\bSi_{\bt}\succ 0\in\R^{d\times d}$. The input sequence and its associated label, i.e., $(\Z,y)$ with $\Z$ denoted in \eqref{def Z app}, 
 are generated as follows:
\begin{itemize}
    \item A task parameter $\bt$ is generated from a Gaussian prior $\bt \sim \mathcal{N}(\bmu,\bSi_\bt)$.
    \item Conditioned on $\bt$, for $i\in[n+1]$, $(\x_i, y_i)$ is generated by $\x_i \sim \mathcal{N}(0, \bSi_{\x})$ and $y_i\sim \mathcal{N}(\x_i^\top\bt, \sigma^2)$.
\end{itemize}
\end{definition}
Here, $\sigma \geq 0$ is the noise level.

In a noisy label setting, the labels $y_i$ in the input sequence $\Z$ can be obtained by
\[
y_i=\x_i^\top\bt+\xi_i, ~\text{where}~\xi_i\sim\Nc(0,\sigma^2), i\in[n+1].
\]
Thus, the labels in the contexts and the label to predict can be obtained by:
\[
\begin{aligned}
\y&=\X\bt+\mtx{\xi},
~\text{where}~ \mtx{\xi}=[\xi_1~\cdots~\xi_n]^\top\in\R^n,\\
y&=\x^\top\bt+\xi_{n+1}.
\end{aligned}
\]
\begin{definition}
[Multi-task ICL]
Consider a multi-task ICL problem with $K$ different tasks. 
Each task generates $(\Z,y)\sim\Dc_k$ following Definition~\ref{def:app_singletask} using shared feature distribution $\x_i \sim \mathcal{N}(0, \bSi_{\x})$, $i\in[n+1]$ but distinct task distributions 
$\bt_k \sim  \mathcal{N}(\bmu_k, \bSi_{\bt_k})$
with mean $\bmu_k$ and covariance $\bSi_{\bt_k}$ for $k \in [K]$.

Additionally, let $\{\pi_k\}_{k=1}^K$ be the probabilities of each task, satisfying $\sum_{k=1}^K \pi_k = 1$ and $\pi_k \geq 0$. 
\end{definition}

We consider a task-aware multi-task ICL setting. Specifically, when a task is selected according to $\pi_k$, its task index $k$ is known. Let $\bar\Dc:=\sum_{k=1}^K\pi_k\Dc_k$ be the mixture of distributions and given sequence model $f:\R^{(n+1)\times(d+1)}\to\R$, we define the multi-task ICL objective as follows:
\begin{align}\label{app:obj}
\Lc(f)&=\E_{(\Z,y)\sim\bar\Dc}[\left(y-f(\Z)\right)^2].
\end{align}

To start with, recap from Definition~\ref{definition:multi-task} where task $k$ has probability $\pi_k$ and its task vector follows distribution $\bt_k\sim\Nc(\bmu_k,\bSi_{\bt_k})$. Following \eqref{eqn:bias} in the main paper, we define the debiased and biased mixed-task covariances (variant with $\bSi_{\x}$ prior)  as follows:
\begin{subequations}
\begin{align}
\text{Debiased:}~&{\debias{\bSbbt}=\bSi_{\x}\sum_{k=1}^K\pi_k\E[(\bt_k-\bmu_k)(\bt_k-\bmu_k)^\top];}\\
\text{Biased:}~\quad &{\bias{\bStbt}
=\bSi_{\x}\sum_{k=1}^K\pi_k\E[\bt_k\bt_k^\top].}
\end{align}
\end{subequations}
Note that they satisfy  $\bSbbt=\bSi_{\x}\sum_{k=1}^K\pi_k\bSi_{\bt_k}$ and  $\bStbt=\bSi_{\x}\sum_{k=1}^K\pi_k(\bSi_{\bt_k}+\bmu_k\bmu_k^\top)$.

We first analyze the plain training setting where no additional task-specific parameters are introduced, and all $K$ tasks are mixed together.

Under Assumption~\ref{assumption:PGD} in the main paper, let the prompt token for task $k$ be $\pb_k = \begin{bmatrix} \pbb_k \\ 1 \end{bmatrix}$. The prediction of a single-layer linear attention model can then be written as:
\begin{align}
f(\Zk)& =\x^\top\W\begin{bmatrix}
    \bar{\pb}_k&\X^\top
\end{bmatrix}\begin{bmatrix}
    1\\\y
\end{bmatrix}\notag\\&=\x^\top \W (\X^\top \y + \pbb_k):=g(\x,\X,\y;\W,\pbb_k).
\end{align}

Note that for the multi-task ICL with task-specific prompting, the optimization object~\eqref{app:obj} can be denoted as:
\[
\begin{aligned}
\Lc(f)
&=\sum_{k=1}^K\pi_k\underbrace{\E_{(\Z,y)\sim\Dc_k}\left[(y- f(\Z^{(k)}))^2\right]}_{\text{Denoted as } \Lc_k(f)}\\
&=\sum_{k=1}^K\pi_k\Lc_k(f)
 =\sum_{k=1}^K\pi_k\E_{(\Z,y)\sim\Dc_k}\left[(y- g(\x,\X,\y;\W,\pbb_k))^2\right]
:=\sum_{k=1}^K\pi_k\Lc_k(\W,\pbb_k),
\end{aligned}
\]
where $\Z=\begin{bmatrix}
    \X^\top&\x\\
    \y^\top&0
\end{bmatrix}^\top\in\R^{(n+1)\times(d+1)},\Zk=\begin{bmatrix}
    \pbb_k&\X^\top&\x\\
    1&\y^\top&0
\end{bmatrix}^\top\in\R^{(n+2)\times (d+1)}.$

Note that the multi-task ICL loss is equivalent to calculating the weighted sum of the task-$k$ ICL losses over all tasks $k \in [K]$.

We begin by deriving the single-task ICL loss $\Lc_k(f)$ for task $k$, and then generalize it to the multi-task ICL loss by taking a weighted sum. Since the derivation is similar for all tasks, the task index $k$ is omitted in the following derivation for simplicity. Unless otherwise specified, $\Lc_k(\W, \pbb_k)$ and $\Lc(\W, \pbb)$ will represent the same meaning. This convention similarly applies to other task-specific parameters, e.g., $\bSi_{\bt} \leftrightarrow \bSi_{\bt_k}$, $\bmu \leftrightarrow \bmu_k$, etc.

The loss on a certain task with a trainable prompt can be determined by:
\begin{align}
\Lc(\W,\pbb)&=\E\left[\left(y-g(\x,\X,\y;\W,\pbb)\right)^2\right]\nn\\
&=\E\left[\left(\x^\top\W\X^\top\y+\x^\top\W\pbb-\x^\top\bt-\xi_{n+1}\right)^2\right]\nn\\
&=\E\left[\left(\x^\top\W\X^\top(\X\bt+\mtx{\xi})+\x^\top\W\pbb-\x^\top\bt-\xi_{n+1}\right)^2\right]\nn\\
&=\E\left[\left(\x^\top\underbrace{\left((\W\X^\top\X-\Iden)(\btb+\bmu)+\W\X^\top\mtx{\xi}+\W\pbb\right)}_{\text{Denoted as a (task-specific) vector }\cb}\right)^2\right]+\sigma^2\nn\\
&=\E\left[\left(\x^\top\cb\right)^2\right]+\sigma^2=\tr{\E[\cb\cb^\top]\bSi_{\x}}+\sigma^2,\label{eqn:loss with c}
\end{align}
where $\btb=\bt-\bmu\sim\Nc(\zerbb,\bSi_{\bt})$ is a centralized variable.

\subsection{Proof of Theorem~\ref{thm:pretrain}}
\begin{theorem} [Plain training] Consider training a single-layer linear attention model in solving multi-task ICL problem with dataset defined in Definition~\ref{definition:multi-task} and model construction as described in Assumption~\ref{assumption:PGD}. Let the optimal solution $\W_{\pt}^\st$ (c.f.~\eqref{eqn:obj pt} in the main paper) and the minimal plain training loss $\Lc_\pt^\st$ as defined in Section~\ref{sec:PGD}. 
Additionally, let $\bStbt$ 
be defined in \eqref{eqn:bt bias} in the main paper and   $\Wb^\st_\pt=\bSi_{\x}\W^\st_\pt$.

Then the solution $\Wb^\star_{\pt}$ and optimal loss $\Lc^\st_\pt$ satisfy
\begin{tcolorbox}[colback=white!5!white,colframe=black!5!black,colback=white!5!white,
                  interior hidden,
                  arc=0pt,
                  boxrule=1pt,
                  boxsep=0pt,
                  left=5pt,
                  right=5pt,
                  top=5pt,]
\[
\scalemath{0.9}{\Wb^\star_{\pt}=\bias{\bStbt}\left((n+1)\bias{\bStbt}+(\tr{\bias{\bStbt}}+\sigma^2)\Iden\right)^{-1},}
\]
\[
\scalemath{0.9}{\Lc^\star_{\pt} 
=\tr{\bias{\bStbt}}+\sigma^2
-n\tr{\Wb^\star_{\pt}\bias{\bStbt}}.}
\]

\end{tcolorbox}
\end{theorem}
\begin{proof}
As previously stated, the following derivation applies to all tasks $k \in [K]$. Therefore, for simplicity, we omit the index $k$ in the notation unless otherwise specified.

In the plain training setting, $\pbb = \zerbb$ for all tasks, and only the attention model, parameterized by $\W$ under Assumption~\ref{assumption:PGD}, is updated. Hence, the loss in \eqref{eqn:loss with c} is:
\[
\begin{aligned}
\Lc(\W,\pbb=\zerbb)&=\tr{\E[\cb\cb^\top]\bSi_{\x}}+\sigma^2,~\text{where}~\cb=\left((\W\X^\top\X-\Iden)(\btb+\bmu)+\W\X^\top\mtx{\xi}\right).
\end{aligned}
\]
The expansion of $\cb\cb^\top$ is (there are $3\times3=9$ terms in total):
\[
\begin{aligned}
\cb\cb^\top=&\left((\W\X^\top\X-\Iden)(\btb+\bmu)+\W\X^\top\mtx{\xi}\right)\left((\W\X^\top\X-\Iden)(\btb+\bmu)+\W\X^\top\mtx{\xi}\right)^\top\\
=&\left((\W\X^\top\X)(\btb+\bmu)-(\btb+\bmu)+\W\X^\top\mtx{\xi}\right)\left((\W\X^\top\X)(\btb+\bmu)-(\btb+\bmu)+\W\X^\top\mtx{\xi}\right)^\top\\
=&\scalemath{0.9}{\left[(\W\X^\top\X)(\btb+\bmu)\right]\left[(\W\X^\top\X)(\btb+\bmu)\right]^\top 
+\left[(\W\X^\top\X)(\btb+\bmu)\right]\left[-(\btb+\bmu)\right]^\top
+\left[(\W\X^\top\X)(\btb+\bmu)\right]\left[\W\X^\top\mtx{\xi}\right]^\top}\\
+&\left[-(\btb+\bmu)\right]\left[(\W\X^\top\X)(\btb+\bmu)\right]^\top 
+\quad\left[-(\btb+\bmu)\right]\left[-(\btb+\bmu)\right]^\top
+\quad\left[-(\btb+\bmu)\right]\left[\W\X^\top\mtx{\xi}\right]^\top\\
+&\left[\W\X^\top\mtx{\xi}\right]\left[(\W\X^\top\X)(\btb+\bmu)\right]^\top 
+\quad\left[\W\X^\top\mtx{\xi}\right]\left[-(\btb+\bmu)\right]^\top
+\quad\left[\W\X^\top\mtx{\xi}\right]\left[\W\X^\top\mtx{\xi}\right]^\top.
\end{aligned}
\]
Take expectation of it,
\[
\begin{aligned}
\E[\cb\cb^\top]&=\left[\W\underbrace{\E[\X^\top\X(\btb+\bmu)(\btb+\bmu)^\top\X^\top\X]}_{\text{Lemma~\ref{lemma:wishart}, denoted as a (task-specific) matrix }\Cb}\W^\top\right]
+\quad\left[-n\W\bSi_{\x}(\bSi_{\bt}+\bmu\bmu^\top)\right]
+\quad 0\\
&+\left[-n(\bSi_{\bt}+\bmu\bmu^\top)\bSi_{\x}\W^\top\right]
+\quad(\bSi_{\bt}+\bmu\bmu^\top)\quad
+\quad 0\quad\\
&+\quad 0\quad
+\quad 0\quad
+\quad \underbrace{n\sigma^2\W\bSi_{\x}\W^\top}_{\text{Lemma~\ref{lemma:noise}}}\\
&=\W(\Cb+n\sigma^2\bSi_{\x})\W^\top-n\W\bSi_{\x}(\bSi_{\bt}+\bmu\bmu^\top)-n(\bSi_{\bt}+\bmu\bmu^\top)\bSi_{\x}\W^\top+(\bSi_{\bt}+\bmu\bmu^\top),
\end{aligned}
\]
where $\Cb=n\tr{\bSi_{\x}(\bSi_{\bt}+\bmu\bmu^\top)}\bSi_{\x}+n(n+1)\bSi_{\x}(\bSi_{\bt}+\bmu\bmu^\top)\bSi_{\x}.$

Substitute back into the loss $\Lc(\W,\pbb=\zerbb)$:
\[
\begin{aligned}
\Lc(\W,\pbb=\zerbb)&=\tr{\W(\Cb+n\sigma^2\bSi_{\x})\W^\top\bSi_{\x}}-2n\tr{\bSi_{\x}\W\bSi_{\x}(\bSi_{\bt}+\bmu\bmu^\top)}+\tr{\bSi_{\x}(\bSi_{\bt}+\bmu\bmu^\top)}+\sigma^2.
\end{aligned}
\]
Use the following definition:
\[
\begin{aligned}
    \text{Debiased:}~&{\debias{\bSbbt}=\bSi_{\x}\sum_{k=1}^K\pi_k\E[(\bt_k-\bmu_k)(\bt_k-\bmu_k)^\top]=\bSi_{\x}\sum_{k=1}^K\pi_k\bSi_{\bt_k};}\\
\text{Biased:}~\quad &{\bias{\bStbt}
=\bSi_{\x}\sum_{k=1}^K\pi_k\E[\bt_k\bt_k^\top]=\bSi_{\x}\sum_{k=1}^K\pi_k(\bSi_{\bt_k}+\bmu_k\bmu_k^\top).}
\end{aligned}
\]
Denote $$\bar\Cb=\sum_{k=1}^K\pi_k\Cb_k=n\tr{\bias{\bStbt}}\bSi_{\x}+n(n+1)\bias{\bStbt}\bSi_{\x}.$$
The weighted sum of the $k$-th task plain training loss over all tasks $k \in [K]$ is:
\[
\begin{aligned}
\Lc_\pt(\W,\Pb=\zerbb)&=\sum_{k=1}^K \pi_k\Lc_k(\W,\pbb_k=\zerbb)\\
&=\tr{\W{(\bar\Cb+n\sigma^2\bSi_{\x})\W^\top\bSi_{\x}}}-2{n}\tr{\W{\bias{\bStbt}\bSi_{\x}}}+\tr{\bias{\bStbt}}+\sigma^2.
\end{aligned}
\]
Take derivative w.r.t. $\W$, using Lemma~\ref{lemma:derivative_2nd}. Since $\Cb$ is symmetric
\[
\frac{\partial\Lc_\pt(\W,\Pb=\zerbb)}{\partial\W}=2\bSi_{\x}\W(\bar\Cb+n\sigma^2\bSi_{\x})-2n\bias{\bStbt}\bSi_{\x}
\]
Let $\Wb^\st_\pt = \bSi_{\x} \W^\st_\pt$, and set the derivative to zero:
\begin{align*}
\Wb^\star_{\pt}&=n\bias{\bStbt}\bSi_{\x}(\bar\Cb+n\sigma^2\bSi_{\x})^{-1}\\
&=\bias{\bStbt}\bSi_{\x}\left((\tr{\bias{\bStbt}}+\sigma^2)\bSi_{\x}+(n+1)\bias{\bStbt}\bSi_{\x}\right)^{-1}\\
&\boxed{=\bias{\bStbt}\left((n+1)\bias{\bStbt}+(\tr{\bias{\bStbt}}+\sigma^2)\Iden\right)^{-1}.}
\end{align*}
Substitute back into $\Lc_\pt(\W, \Pb = \zerbb)$:
\[\boxed{\Lc^\star_{\pt} 
=\tr{\bias{\bStbt}}+\sigma^2
-n\tr{\Wb^\star_{\pt}\bias{\bStbt}}.}\]
\end{proof}

\subsection{Proof of Theorem~\ref{thm:pretrain-finetune}}
\begin{theorem} [Fine-tuning]Suppose a pretrained model as described in Theorem~\ref{thm:pretrain} is given with $\W^\st_\pt$ being its optimal solution. Consider fine-tuning this model with task-specific prompts as defined in Definition~\ref{def:task-specific param}, and let the optimal prompt matrix $\Pb^\st_\ft$ (c.f.~\eqref{eqn:obj ft} in the main paper) and the minimal fine-tuning loss $\Lc^\st_\ft$ be defined in Section~\ref{sec:PGD}. Additionally, let $\bSbbt,\bStbt$ be defined in \eqref{eqn:bt bias} in the main paper and $\Wb^\st_\pt=\bSi_{\x}\W^\st_\pt$,  and define the mean matrix
\[
\M_\bmu=[\bmu_1~\cdots~\bmu_K]^\top\in\R^{K\times d}.
\]
Then the solution $\Pb^\st_\ft$ and optimal loss $\Lc^\st_\ft$ satisfy 
\begin{tcolorbox}[colback=white!5!white,colframe=black!5!black,colback=white!5!white,
                  interior hidden,
                  arc=0pt,
                  boxrule=1pt,
                  boxsep=0pt,
                  left=5pt,
                  right=5pt,
                  top=5pt,]
\[
\scalemath{0.9}{\Pb^\st_\ft = \M_\bmu\left((\Wb^\star_{\pt})^{-1} - n\Iden\right)\bSi_{\x},}
\]
\[
\scalemath{0.9}{\Lc^\star_\ft
=\Lc^\star_{\pt}-\tr{(\bias{\bStbt}-\debias{\bSbbt})(n\Wb^{\star}_{\pt}-\Iden)^\top(n\Wb^{\star}_{\pt}-\Iden)}.}
\]
\end{tcolorbox}
\end{theorem}
\begin{proof}
As previously stated, the following derivation applies to all tasks $k \in [K]$. Therefore, for simplicity, we omit the index $k$ in the notation unless otherwise specified.

\textbf{1. Determining the optimal task-specific prompts}

In the fine-tuning setting, the attention model is pretrained and parameterized by a fixed $\W = \W^\st_\pt$, and only the task-specific prompts are fine-tuned. Then recap from \eqref{eqn:loss with c}:
\[
\begin{aligned}
&\Lc(\W,\pbb) = \tr{\E[\cb\cb^\top]\bSi_{\x}} + \sigma^2, \\
&\text{where}\quad\cb = \left((\W \X^\top \X - \Iden)(\btb + \bmu) + \W \X^\top \mtx{\xi} + \W \pbb \right).
\end{aligned}
\]
The optimal task-specific prompt is determined by taking the derivative and setting it to zero:
\[
\begin{aligned}
\frac{\partial \Lc(\W,\pbb)}{\partial \pbb} &= 2 \E\left[\frac{\partial \cb}{\partial \pbb}^\top \bSi_{\x} \cb \right] \\
&= 2 \W^\top \bSi_{\x} \E[\cb] = 2 \W^\top \bSi_{\x} \left[(n \W \bSi_{\x} - \Iden) \bmu + \W \pbb \right] = \zerbb, \\
\Rightarrow \quad \pbb^\st &= \left(\W^{-1} - n \bSi_{\x} \right) \bmu,
\end{aligned}
\]
which is equivalent to (by substituting $\W = \W^\st_\pt$):
\[
\boxed{\Pb^\st_\ft = \M_\bmu \left((\Wb^\star_{\pt})^{-1} - n \bSi_{\x}\right).}
\]

\textbf{2. Determining the fine-tuning loss}

Substituting back $\pbb^\st = \left(\W^{-1} - n \bSi_{\x} \right) \bmu$:
\[
\begin{aligned}
\Lc(\W,\pbb) &= \tr{\E[\cb\cb^\top]\bSi_{\x}} + \sigma^2, \\
\text{where}~~\cb &= \left((\W \X^\top \X - \Iden)(\btb + \bmu) + \W \X^\top \mtx{\xi} + \W \pbb \right)\\
&=\W \X^\top \X(\btb + \bmu)-\btb+\W \X^\top \mtx{\xi}-n\W\bSi_{\x}\bmu.
\end{aligned}
\]
The expansion of $\cb\cb^\top$ is (there are $4\times4=16$ terms in total):
\[
\begin{aligned}
\cb\cb^\top =& \left[\W \X^\top \X(\btb + \bmu)-\btb+\W \X^\top \mtx{\xi}-n\W\bSi_{\x}\bmu\right]\left[\W \X^\top \X(\btb + \bmu)-\btb+\W \X^\top \mtx{\xi}-n\W\bSi_{\x}\bmu\right]^\top\\
=&\scalemath{0.75}{\left[\W \X^\top \X(\btb + \bmu)\right]\left[\W \X^\top \X(\btb + \bmu)\right]^\top
+\left[\W \X^\top \X(\btb + \bmu)\right]\left[-\btb\right]^\top
+\left[\W \X^\top \X(\btb + \bmu)\right]\left[\W \X^\top \mtx{\xi}\right]^\top
+\left[\W \X^\top \X(\btb + \bmu)\right]\left[-n\W\bSi_{\x}\bmu\right]^\top}\\
+&\scalemath{0.8}{\quad\left[-\btb\right]\left[\W \X^\top \X(\btb + \bmu)\right]^\top\quad
+\quad\quad\left[-\btb\right]\left[-\btb\right]^\top\quad
+\quad\quad\left[-\btb\right]\left[\W \X^\top \mtx{\xi}\right]^\top\quad
+\quad\quad\left[-\btb\right]\left[-n\W\bSi_{\x}\bmu\right]^\top}\quad\\
+&\scalemath{0.8}{\left[\W \X^\top \mtx{\xi}\right]\left[\W \X^\top \X(\btb + \bmu)\right]^\top
+\left[\W \X^\top \mtx{\xi}\right]\left[-\btb\right]^\top\quad
+\left[\W \X^\top \mtx{\xi}\right]\left[\W \X^\top \mtx{\xi}\right]^\top\quad
+\left[\W \X^\top \mtx{\xi}\right]\left[-n\W\bSi_{\x}\bmu\right]^\top}\\
+&\scalemath{0.77}{\left[-n\W\bSi_{\x}\bmu\right]\left[\W \X^\top \X(\btb + \bmu)\right]^\top
+\left[-n\W\bSi_{\x}\bmu\right]\left[-\btb\right]^\top
+\left[-n\W\bSi_{\x}\bmu\right]\left[\W \X^\top \mtx{\xi}\right]^\top
+\left[-n\W\bSi_{\x}\bmu\right]\left[-n\W\bSi_{\x}\bmu\right]^\top}.
\end{aligned}
\]
Take expectation of it,
\[
\begin{aligned}
\E[\cb\cb^\top]=&\left[\W\underbrace{\E[\X^\top\X(\btb+\bmu)(\btb+\bmu)^\top\X^\top\X]}_{\text{Lemma~\ref{lemma:wishart}, denoted as a (task-specific) matrix }\Cb}\W^\top\right]
+\left[-n\W\bSi_{\x}\bSi_{\bt}\right]
+\quad 0\quad
+\left[-n^2\W\bSi_{\x}\bmu\bmu^\top\bSi_{\x}\W^\top\right]\\
+&\left[-n\bSi_{\bt}\bSi_{\x}\W^\top\right]\quad
+\quad\bSi_{\bt}\quad
+\quad 0\quad
+\quad 0\quad\\
+&\quad 0\quad
+\quad 0\quad
+\underbrace{n\sigma^2\W\bSi_{\x}\W^\top}_{\text{Lemma~\ref{lemma:noise}}}
+\quad 0\\
+&\left[-n^2\W\bSi_{\x}\bmu\bmu^\top\bSi_{\x}\W^\top\right]
+\quad 0\quad
+\quad 0\quad
+\quad\left[n^2\W\bSi_{\x}\bmu\bmu^\top\bSi_{\x}\W^\top\right]\\
=&\W(\Cb+n\sigma^2\bSi_{\x}-n^2\bSi_{\x}\bmu\bmu^\top\bSi_{\x})\W^\top-n\W\bSi_{\x}\bSi_{\bt}-n\bSi_{\bt}\bSi_{\x}\W^\top+\bSi_{\bt}
\end{aligned}
\]
where $\Cb=n\tr{\bSi_{\x}(\bSi_{\bt}+\bmu\bmu^\top)}\bSi_{\x}+n(n+1)\bSi_{\x}(\bSi_{\bt}+\bmu\bmu^\top)\bSi_{\x}.$

Substitute back into the loss $\Lc(\W,\pbb^\st(\W))$:
\[
\begin{aligned}
\Lc(\W,\pbb^\st(\W))&=\tr{\E[\cb\cb^\top]\bSi_{\x}} + \sigma^2\\
&=\tr{\W(\Cb+n\sigma^2\bSi_{\x}-n^2\bSi_{\x}\bmu\bmu^\top\bSi_{\x})\W^{\top}\bSi_{\x}}-2n\tr{\W\bSi_{\x}\bSibt\bSi_{\x}}+\tr{\bSibt\bSi_{\x}}+\sigma^2
\end{aligned}
\]
Use the following definition:
\[
\begin{aligned}
    \text{Debiased:}~&{\debias{\bSbbt}=\bSi_{\x}\sum_{k=1}^K\pi_k\E[(\bt_k-\bmu_k)(\bt_k-\bmu_k)^\top]=\bSi_{\x}\sum_{k=1}^K\pi_k\bSi_{\bt_k};}\\
\text{Biased:}~\quad &{\bias{\bStbt}
=\bSi_{\x}\sum_{k=1}^K\pi_k\E[\bt_k\bt_k^\top]=\bSi_{\x}\sum_{k=1}^K\pi_k(\bSi_{\bt_k}+\bmu_k\bmu_k^\top).}
\end{aligned}
\]

Denote: 
$$\bar\Cb=\sum_{k=1}^K\pi_k\Cb_k=n\tr{\bias{\bStbt}}\bSi_{\x}+n(n+1)\bias{\bStbt}\bSi_{\x}.$$
The weighted sum of the $k$-th task fine-tuning loss over all tasks $k \in [K]$ is:
\[
\begin{aligned}
\Lc_\ft(\W,\Pb_\ft^\st(\W))=\tr{\W\big(\bar\Cb+n\sigma^2\bSi_{\x}-n^2 (\bias{\bStbt}-\debias{\bSbbt})\bSi_{\x}\big)\W^{\top}\bSi_{\x}}-2n\tr{\W\debias{\bSbbt}\bSi_{\x}}+\tr{\debias{\bSbbt}}+\sigma^2
\end{aligned}
\]
Note that for plain training,
\[
\Lc_\pt(\W,\Pb=\zerbb)=\tr{\W(\bar\Cb+n\sigma^2\bSi_{\x})\W^\top\bSi_{\x}}-2n\tr{\W\bias{\bStbt}\bSi_{\x}}+\tr{\bias{\bStbt}}+\sigma^2,
\]
and their optimal loss share the same attention weight parameterization $\W=\W^\st_\pt$. Let $\Wb^\st_\pt=\bSi_{\x}\W^\st_\pt,$
\[
\begin{aligned}
\mathcal{L}^\star_\ft &= \mathcal{L}^\star_{\pt}-\tr{\bias{\bStbt}-\debias{\bSbbt}}
+2n\tr{\Wb^\star_{\pt}(\bias{\bStbt}-\debias{\bSbbt})}-n^2\tr{\Wb^\star_{\pt}(\bias{\bStbt}-\debias{\bSbbt})\Wb^{\star\top}_{\pt}}\\
&\boxed{=\mathcal{L}^\star_{\pt}-\tr{(\bias{\bStbt}-\debias{\bSbbt})(n\Wb^{\star}_{\pt}-\Iden)^\top(n\Wb^{\star}_{\pt}-\Iden)}.}
\end{aligned}
\]

\end{proof}

\subsection{Proof of Theorem~\ref{thm:joint-training}}
\begin{theorem}[Joint training]Consider training a single-layer linear attention model in solving multi-task ICL problem with dataset defined in Definition~\ref{definition:multi-task} and model construction as described in Assumption~\ref{assumption:PGD} in the main paper. Let $\W_{\jt}^\st,\Pb^\st_\jt$ (c.f.~\eqref{eqn:obj jt} in the main paper) be the optimal solutions and $\Lc_\jt^\st$ is the optimal joint training loss defined in Section~\ref{sec:PGD}. 
Additionally, let $\bSbbt,\bStbt,\M_\bmu$ follow the same definitions as in Theorem~\ref{thm:pretrain-finetune} and define $\Wb^\st_\jt=\bSi_{\x}\W^\st_\jt$. Then the solution $(\W_{\jt}^\st,\Pb^\st_\jt)$ and optimal loss $\Lc^\st_\jt$ satisfy 
\begin{tcolorbox}[colback=white!5!white,colframe=black!5!black,colback=white!5!white,
                  interior hidden,
                  arc=0pt,
                  boxrule=1pt,
                  boxsep=0pt,
                  left=5pt,
                  right=5pt,
                  top=5pt,]
\[
\scalemath{0.9}{\Wb^\star_{\jt} = {\debias{\bSbbt}}
\left((n+1)\debias{\bSbbt}+ (\tr{\bias{\bStbt}}+\sigma^2) \Iden+\bSi_{\x}\sum_{k=1}^K\pi_k\bmu_k\bmu_k^\top\right)^{-1},}
\]\[
\scalemath{0.9}{\Pb_\jt^\star = \M_\bmu\left((\Wb^\star_{\jt})^{-1} - n\Iden\right)\bSi_{\x},}
\]
\[
\scalemath{0.9}{\Lc^\star_{\jt} =\tr{\debias{\bSbbt}}+\sigma^2-n\tr{\Wb^\star_{\jt}{\debias{\bSbbt}}}.}
\]
\end{tcolorbox}
Here, $\bSi_{\x}\sum_{k=1}^K\pi_k\bmu_k\bmu_k^\top\sim\order{1}$ is a $d\times d$-sized constant matrix. 
\end{theorem}
\begin{proof}
As previously stated, the following derivation applies to all tasks $k \in [K]$. Therefore, for simplicity, we omit the index $k$ in the notation unless otherwise specified.

\textbf{1. Determining the optimal task-specific prompts}

In the joint training setting, the attention model is pretrained and parameterized by a trainable $\W$, and the task-specific prompts are fine-tuned accordingly (see \eqref{eqn:loss with c}):
\[
\begin{aligned}
&\Lc(\W,\pbb) = \tr{\E[\cb\cb^\top]\bSi_{\x}} + \sigma^2, \\
&\text{where}\quad\cb = (\W \X^\top \X - \Iden)(\btb + \bmu) + \W \X^\top \mtx{\xi} + \W \pbb .
\end{aligned}
\]
The optimal task-specific prompt is determined by taking the derivative and setting it to zero:
\[
\begin{aligned}
\frac{\partial \Lc(\W,\pbb)}{\partial \pbb} &= 2 \E\left[\frac{\partial \cb}{\partial \pbb}^\top \bSi_{\x} \cb \right] \\
&= 2 \W^\top \bSi_{\x} \E[\cb] = 2 \W^\top \bSi_{\x} \left[(n \W \bSi_{\x} - \Iden) \bmu + \W \pbb \right] = \zerbb, \\
\Rightarrow \quad \pbb^\st &= \left(\W^{-1} - n \bSi_{\x} \right) \bmu,
\end{aligned}
\]
which is equivalent to:
\[
\boxed{\Pb_\jt^\star = \M_\bmu\left((\Wb^\star_{\jt})^{-1} - n\Iden\right)\bSi_{\x}.}
\]

\textbf{2. Determining the fine-tuning loss}

It is worth noting that fine-tuning and joint training share a functional relationship between the tuned prompts $\pbb^\st$ and the current attention model parameterization $\W$. Thus, substituting the tuned prompt back into the joint training loss will result in the same expression as the fine-tuning loss:

\[
\begin{aligned}
\Lc(\W,\pbb^\st(\W))&=\tr{\E[\cb\cb^\top]\bSi_{\x}} + \sigma^2\\
&=\tr{\W(\Cb+n\sigma^2\bSi_{\x}-n^2\bSi_{\x}\bmu\bmu^\top\bSi_{\x})\W^{\top}\bSi_{\x}}-2n\tr{\W\bSi_{\x}\bSibt\bSi_{\x}}+\tr{\bSibt\bSi_{\x}}
\end{aligned}
\]
Use the following definition:
\[
\begin{aligned}
    \text{Debiased:}~&{\debias{\bSbbt}=\bSi_{\x}\sum_{k=1}^K\pi_k\E[(\bt_k-\bmu_k)(\bt_k-\bmu_k)^\top]=\bSi_{\x}\sum_{k=1}^K\pi_k\bSi_{\bt_k};}\\
\text{Biased:}~\quad &{\bias{\bStbt}
=\bSi_{\x}\sum_{k=1}^K\pi_k\E[\bt_k\bt_k^\top]=\bSi_{\x}\sum_{k=1}^K\pi_k(\bSi_{\bt_k}+\bmu_k\bmu_k^\top).}
\end{aligned}
\]
Denote: $$\bar\Cb=\sum_{k=1}^K\pi_k\Cb_k=n\tr{\bias{\bStbt}}\bSi_{\x}+n(n+1)\bias{\bStbt}\bSi_{\x}.$$
The weighted sum of the $k$-th task fine-tuning loss over all tasks $k \in [K]$ is:
\[
\begin{aligned}
\Lc_\jt(\W,\Pb_\jt^\st(\W))=\tr{\W\big(\bar\Cb+n\sigma^2\bSi_{\x}-n^2 (\bias{\bStbt}-\debias{\bSbbt})\bSi_{\x}\big)\W^{\top}\bSi_{\x}}-2n\tr{\W\debias{\bSbbt}\bSi_{\x}}+\tr{\debias{\bSbbt}}+\sigma^2.
\end{aligned}
\]
In joint training, the attention model parameterization $\W$ is no longer fixed (as it is in the fine-tuning setting), but is instead optimized. Due to the functional relationship between the tuned prompts and parameterization $\pbb^\st=\pbb^\st(\W)$, they will be optimized jointly until reaching their optimal values.

Take derivative w.r.t. $\W$, using Lemma~\ref{lemma:derivative_2nd}:
\[
\frac{\partial\Lc_\jt(\W,\Pb_\jt^\st(\W))}{\partial\W}=2\bSi_{\x}\W\big(\bar\Cb+n\sigma^2\bSi_{\x}-n^2 (\bias{\bStbt}-\debias{\bSbbt})\bSi_{\x}\big)-2n\debias{\bSbbt}\bSi_{\x}
\]
Let $\Wb^\st_\jt = \bSi_{\x} \W^\st_\jt$, and set the derivative to zero (note that $\sum_{k=1}^K\bmu_k\bmu_k^\top\sim\order{1}$):
\[
\begin{aligned}
\Wb^\star_{\jt} &=n\debias{\bSbbt}\bSi_{\x}\left(\bar\Cb+n\sigma^2\bSi_{\x}-n^2(\bias{\bStbt}-\debias{\bSbbt})\bSi_{\x})\right)^{-1}\\
&=\debias{\bSbbt}\bSi_{\x}\left((\tr{\bias{\bStbt}}+\sigma^2)\bSi_{\x}-n(\bias{\bStbt}-\debias{\bSbbt})\bSi_{\x}+(n+1)\bias{\bStbt}\bSi_{\x}\right)^{-1}\\
&=\debias{\bSbbt}\left((\tr{\bias{\bStbt}}+\sigma^2)\Iden+(n+1)\debias{\bSbbt}+(\bias{\bStbt}-\debias{\bSbbt})\right)^{-1}\\
&\boxed{= {\debias{\bSbbt}}
\left((n+1)\debias{\bSbbt}+ (\tr{\bias{\bStbt}}+\sigma^2) \Iden+\bSi_{\x}\sum_{k=1}^K\pi_k\bmu_k\bmu_k^\top\right)^{-1}.}
\end{aligned}
\]
Substitute back into $\Lc_\jt(\W,\Pb^\st(\W))$:
\[\boxed{\Lc^\star_{\jt} 
=\tr{\debias{\bSbbt}}+\sigma^2
-n\tr{\Wb^\star_{\jt}\debias{\bSbbt}}.}\]
\end{proof}

\subsection{Proof of Corollary~\ref{corollary:ordering}}
\begin{corollary}
Let $\Lc^*_{\pt}$, $\Lc^*_{\ft}$, and $\Lc^*_{\jt}$ denote the optimal losses for plain training, fine-tuning, and joint training, as described in Theorems 1, 2 and 3, respectively. These losses satisfy:
\begin{align}
\Lc^*_{\jt} \leq \Lc^*_{\ft} \leq \Lc^*_{\pt}.
\end{align}
The equalities hold if and only if $\bSb_\beta = \bSp_\beta$ (c.f. (14)), which occurs when all task means $\bmu_k = \zerbb$ for $k \in [K]$. Furthermore, the loss gaps satisfy the following:
\begin{enumerate}
    \item The loss gaps scale quadratically with task mean:
    ${
       \Lc^*_{\pt} - \Lc^*_{\ft} \sim \mathcal{O}\left(\frac{1}{n^2}\right)\|\Delta\|_F,\text{ } 
       \Lc^*_{\ft} - \Lc^*_{\jt} \sim \mathcal{O}\left(\frac{1}{n}\right)\|\Delta\|_F},
       $
    where $\Delta := \bias{\bStbt}-\debias{\bSbbt} = \bSi_x\sum_{k=1}^K \pi_k\bmu_k\bmu_k^\top$.
    
    \item The ratio between gaps is: $\frac{\Lc^*_{\pt} - \Lc^*_{\ft}}{ \Lc^*_{\ft}- \Lc^*_{\jt} } \sim \mathcal{O}\left(\frac{1}{n}\right)$,
    indicating that fine-tuning provides most of the benefit in few-shot regimes (small $n$), while joint training benefits more for larger $n$.
\end{enumerate}
\end{corollary}
\begin{proof}
\textbf{1. $\Lc^\st_\ft \leq \Lc^\st_\pt$:}

From Theorem~\ref{thm:pretrain-finetune}, we have:
\[
\Lc^\st_\ft=\Lc^\star_{\pt}-\tr{(\bias{\bStbt}-\debias{\bSbbt})(n\Wb^{\star}_{\pt}-\Iden)^\top(n\Wb^{\star}_{\pt}-\Iden)}.
\]
Use the following definition:
\[
\begin{aligned}
    \text{Debiased:}~&{\debias{\bSbbt}=\bSi_{\x}\sum_{k=1}^K\pi_k\E[(\bt_k-\bmu_k)(\bt_k-\bmu_k)^\top]=\bSi_{\x}\sum_{k=1}^K\pi_k\bSi_{\bt_k};}\\
\text{Biased:}~\quad &{\bias{\bStbt}
=\bSi_{\x}\sum_{k=1}^K\pi_k\E[\bt_k\bt_k^\top]=\bSi_{\x}\sum_{k=1}^K\pi_k(\bSi_{\bt_k}+\bmu_k\bmu_k^\top).}
\end{aligned}
\]
We define an auxiliray variable:
\[
\bSp_\bt - \bSb_\bt = \bSi_x\sum_{k=1}^K \pi_k\bmu_k\bmu_k^\top = \Delta
\]
It can be seen that $(\bias{\bStbt}-\debias{\bSbbt})=\bSi_{\x}(\sum_{k=1}^K\pi_k\bmu_k\bmu_k^\top)\succeq \zerbb, (n\Wb^{\star}_{\pt}-\Iden)^\top(n\Wb^{\star}_{\pt}-\Iden)\succ \zerbb,$ which leads to
\[
\tr{(\bias{\bStbt}-\debias{\bSbbt})(n\Wb^{\star}_{\pt}-\Iden)^\top(n\Wb^{\star}_{\pt}-\Iden)}\geq0\Rightarrow \Lc^\st_\ft \leq \Lc^\st_\pt.
\]
The equality holds if and only if $\boxed{ \bmu_k=\zerbb,k\in[K]\iff\bSbbt = \bStbt.}$

The loss gap between $\Lc^\st_\ft$ and $ \Lc^\st_\pt$ can be written as:
\[
\Lc_{\pt}^\st-\Lc_{\ft}^\st = \tr{\Delta(n\Wb_{\pt}^\st - \Ib)^\top(n\Wb_{\pt}^\st - \Ib)}
\]
To analyze the asymptotic behavior of the gap, we need to analyze the asymptotic behavior of:
\[
\Wb_{\pt}^\st = \bSp_\bt((n+1)\bSp_\bt + \tr{\bSp_\bt}\Ib)^{-1}.
\]

First, for large $n$, we can factor out $n$ from the inverse term:
\[
\Wb_{\pt}^\st = \frac{1}{n}\bSp_\bt\left(\bSp_\bt\left(1 + \frac{1}{n}\right) + \frac{\tr{\bSp_\bt}}{n}\Ib\right)^{-1}
\]

Using a matrix Taylor expansion, with $\A = \bSp_\bt$ and $\B = \frac{\tr{\bSp_\bt}}{n}\Ib + \frac{1}{n}\bSp_\bt$:
\[
(\A + \B)^{-1} = \A^{-1} - \A^{-1}\B\A^{-1} + \mathcal{O}(\|\B\|^2)
\]
Applying this to our expression:
\[
\left(\bSp_\bt\left(1 + \frac{1}{n}\right) + \frac{\tr{\bSp_\bt}}{n}\Ib\right)^{-1} = 
\bSp_\bt^{-1} - \frac{1}{n}\bSp_\bt^{-1}\left(\bSp_\bt + \tr{\bSp_\bt}\Ib\right)\bSp_\bt^{-1} + \mathcal{O}\left(\frac{1}{n^2}\right)
\]
Therefore:
\[
\Wb_{\pt}^\st = \frac{1}{n}\Ib - \frac{1}{n^2}\left(\Ib + \tr{\bSp_\bt}\bSp_\bt^{-1}\right) + \mathcal{O}\left(\frac{1}{n^3}\right)
\]
The omitted $\order{\frac{1}{n^3}}$ terms include higher-order expansion terms from the matrix Taylor series. These terms involve powers of $\bSp_\bt$ and its inverse. They grow increasingly small as $n$ increases and don't affect the dominant $\order{\frac{1}{n^2}}$ behavior of the loss gap.

Using the result above:
\[
n\Wb_{\pt}^\st - \Ib = n\left(\frac{1}{n}\Ib - \frac{1}{n^2}\left(\Ib + \tr{\bSp_\bt}\bSp_\bt^{-1}\right) + \mathcal{O}\left(\frac{1}{n^3}\right)\right) - \Ib
\]

\[
n\Wb_{\pt}^\st - \Ib = \Ib - \frac{1}{n}\left(\Ib + \tr{\bSp_\bt}\bSp_\bt^{-1}\right) + \mathcal{O}\left(\frac{1}{n^2}\right) - \Ib
\]

\[
n\Wb_{\pt}^\st - \Ib = -\frac{1}{n}\left(\Ib + \tr{\bSp_\bt}\bSp_\bt^{-1}\right) + \mathcal{O}\left(\frac{1}{n^2}\right)
\]
The omitted terms at order $\order{\frac{1}{n^2}}$ include additional matrix products that become negligible for large $n$.

Therefore:
\[
\|n\Wb_{\pt}^\st - \Ib\|_F \sim \mathcal{O}\left(\frac{1}{n}\right)
\]
Now, substituting this into the expression for the gap:
\[
\Lc_{\pt}^\st-\Lc_{\ft}^\st  = \tr{\Delta(n\Wb_{\pt}^\st - \Ib)^\top(n\Wb_{\pt}^\st - \Ib)}
\]

\[
\Lc_{\pt}^\st-\Lc_{\ft}^\st \sim  \tr{\Delta \cdot \mathcal{O}\left(\frac{1}{n^2}\right)}
\]

\[
\Lc_{\pt}^\st-\Lc_{\ft}^\st  \sim \mathcal{O}\left(\frac{1}{n^2}\right)\|\Delta\|_F
\]

Given that $\|\Delta\|_F \leq M^2\sum_{k=1}^K \pi_k^2 \cdot \lambda_{\max}(\bSi_x)$, we have:
\[\boxed{
\Lc_{\pt}^\st-\Lc_{\ft}^\st  \sim \mathcal{O}\left(\frac{1}{n^2}\right)M^2\sum_{k=1}^K \pi_k^2 \cdot \lambda_{\max}(\bSi_x)}
\]

\textbf{2. $\Lc^\st_\jt \leq \Lc^\st_\ft$:}

From the proof of Theorem~\ref{thm:joint-training}, it can be seen that joint training and fine-tuning shares a functional relationship between $\Lc$ and $\W$:
\[
\Lc_\jt(\W,\Pb_\jt^\st(\W))=\Lc_\ft(\W,\Pb_\ft^\st(\W))
\]
However, the only minimizer of this funtion is derived from $\frac{\partial\Lc_\jt(\W,\Pb^\st(\W))}{\partial\W}=0\Rightarrow \W^\st_\jt,$ which leads to:
\[
\Lc^\st_\jt \leq \Lc^\st_\ft.
\]
The equality holds if and only if $\boxed{\W^\st_\pt=\W^\st_\jt\iff \bmu_k=\zerbb,k\in[K]\iff\bSbbt = \bStbt.}$

Recall that $\Wb_{\jt}^\st$ is given by:
\[
\Wb_{\jt}^\st = \bSb_\bt((n+1)\bSb_\bt + \tr{\bSp_\bt}\Ib + \bSi_x\sum_{k=1}^K \pi_k\bmu_k\bmu_k^\top)^{-1}
\]

Note that we denote $\Delta = \bSi_x\sum_{k=1}^K \pi_k\bmu_k\bmu_k^\top$, so:
\begin{align*}
\Wb_{\jt}^\st &= \bSb_\bt((n+1)\bSb_\bt + \tr{\bSb_\bt + \Delta}\Ib + \Delta)^{-1}\\
&= \bSb_\bt((n+1)\bSb_\bt + \tr{\bSb_\bt}\Ib + \tr{\Delta}\Ib + \Delta)^{-1}
\end{align*}

First, we rewrite this for large $n$:
\[
\Wb_{\jt}^\st = \frac{1}{n}\bSb_\bt\left(\bSb_\bt\left(1 + \frac{1}{n}\right) + \frac{\tr{\bSb_\bt}}{n}\Ib + \frac{\tr{\Delta}}{n}\Ib + \frac{\Delta}{n}\right)^{-1}
\]

Let $\A = \bSb_\bt$ and $\B = \frac{1}{n}\bSb_\bt + \frac{\tr{\bSb_\bt}}{n}\Ib + \frac{\tr{\Delta}}{n}\Ib + \frac{\Delta}{n}$.

Using the matrix Taylor expansion for $(\A + \B)^{-1}$:
\[
(\A + \B)^{-1} = \A^{-1} - \A^{-1}\B\A^{-1} + \A^{-1}\B\A^{-1}\B\A^{-1} + \mathcal{O}(\|\B\|^3)
\]

Applying this to our expression and noting that $\|\B\| = \mathcal{O}(1/n)$:
\[
\left(\bSb_\bt\left(1 + \frac{1}{n}\right) + \frac{\tr{\bSb_\bt}}{n}\Ib + \frac{\tr{\Delta}}{n}\Ib + \frac{\Delta}{n}\right)^{-1} = \\
\bSb_\bt^{-1} - \frac{1}{n}\bSb_\bt^{-1}\left(\bSb_\bt + \tr{\bSb_\bt}\Ib + \tr{\Delta}\Ib + \Delta\right)\bSb_\bt^{-1} + \mathcal{O}\left(\frac{1}{n^2}\right)
\]

Therefore:
\[
\Wb_{\jt}^\st = \frac{1}{n}\Ib - \frac{1}{n^2}\left(\Ib + \tr{\bSb_\bt}\bSb_\bt^{-1} + \tr{\Delta}\bSb_\bt^{-1} + \bSb_\bt^{-1}\Delta\right) + \mathcal{O}\left(\frac{1}{n^3}\right)
\]
The omitted $\order{\frac{1}{n^3}}$ terms include higher-order matrix products involving powers of $\bSb_\bt$, $\Delta$, and their inverses. These become negligible as $n$ grows.

Using the result above:
\begin{align*}
n\Wb_{\jt}^\st - \Ib &= n\left(\frac{1}{n}\Ib - \frac{1}{n^2}\left(\Ib + \tr{\bSb_\bt}\bSb_\bt^{-1} + \tr{\Delta}\bSb_\bt^{-1} + \bSb_\bt^{-1}\Delta\right) + \mathcal{O}\left(\frac{1}{n^3}\right)\right) - \Ib\\
&= -\frac{1}{n}\left(\Ib + \tr{\bSb_\bt}\bSb_\bt^{-1} + \tr{\Delta}\bSb_\bt^{-1} + \bSb_\bt^{-1}\Delta\right) + \mathcal{O}\left(\frac{1}{n^2}\right)
\end{align*}

Therefore:
\[
\|n\Wb_{\jt}^\st - \Ib\|_F = \mathcal{O}\left(\frac{1}{n}\right)
\]
Recall that:
\[
\Wb_{\pt}^\st = \frac{1}{n}\Ib - \frac{1}{n^2}\left(\Ib + \tr{\bSp_\bt}\bSp_\bt^{-1}\right) + \mathcal{O}\left(\frac{1}{n^3}\right)
\]

\[
\Wb_{\jt}^\st = \frac{1}{n}\Ib - \frac{1}{n^2}\left(\Ib + \tr{\bSb_\bt}\bSb_\bt^{-1} + \tr{\Delta}\bSb_\bt^{-1} + \bSb_\bt^{-1}\Delta\right) + \mathcal{O}\left(\frac{1}{n^3}\right)
\]

The leading terms $(1/n)\Ib$ cancel, and the difference appears in the second-order terms:
\[
\Wb_{\pt}^\st - \Wb_{\jt}^\st = \frac{1}{n^2}\Cb + \mathcal{O}\left(\frac{1}{n^3}\right)
\]

Where $\Cb$ is a matrix that depends on $\bSb_\bt$ and $\Delta$. Therefore:
\[
\|\Wb_{\pt}^\st - \Wb_{\jt}^\st\|_F \sim \mathcal{O}\left(\frac{1}{n^2}\right)\|\Delta\|_F
\]

Starting from the loss function definition:
\[
\Lc(\W) = \tr{\bSb_\bt} - n\tr{\W\bSb_\bt}
\]

The loss gap becomes:
\[
\Lc_{\ft}^\st-\Lc_{\jt}^\st = n\|\tr{(\Wb_{\pt}^\st - \Wb_{\jt}^\st)\bSb_\bt}\|
\]

This gives us:
\[\boxed{
\Lc_{\ft}^\st-\Lc_{\jt}^\st \sim \mathcal{O}\left(\frac{1}{n}\right)M^2\sum_{k=1}^K \pi_k^2 \cdot \lambda_{\max}(\bSi_x)}
\]

\textbf{3. Comparative Analysis}
\begin{enumerate}
\item Plain Training vs Fine-tuning:
\[
\Lc_{\pt}^\st-\Lc_{\ft}^\st  \sim \mathcal{O}\left(\frac{1}{n^2}\right)M^2\sum_{k=1}^K \pi_k^2 \cdot \lambda_{\max}(\bSi_x)
\]

\item Fine-tuning vs Joint Training:
\[
\Lc_{\ft}^\st-\Lc_{\jt}^\st \sim \mathcal{O}\left(\frac{1}{n}\right)M^2\sum_{k=1}^K \pi_k^2 \cdot \lambda_{\max}(\bSi_x)
\]

\item Ratio of gaps:
\[
\frac{\Lc_{\pt}^\st-\Lc_{\ft}^\st }{\Lc_{\ft}^\st-\Lc_{\jt}^\st} \sim \mathcal{O}\left(\frac{1}{n}\right)
\]
\end{enumerate}
This aligns with our experimental results in Section 6, indicating that for few-shot multi-task in-context learning settings (e.g., when $n=1$), fine-tuning task-specific prompts alone is sufficiently effective at improving performance. However, for many-shot (large $n$) settings, joint training of both the attention weight and task-specific prompts is necessary to achieve further performance improvements.
\end{proof}

%% file: app/Appendix/proof_fully_decoupled.tex
\subsection{Proof of Proposition~\ref{prop h}}
\begin{proposition}
Consider the multi-task ICL data as described in Definition~\ref{definition:multi-task} and let $\tilde\Lc_\att^\st$ and $\tilde\Lc_\pgd^\st$ be the optimal linear attention and debiased preconditioned gradient descent losses as presented in \eqref{eqn:obj att h} and \eqref{eqn:obj pgd h} in the main paper, respectively. Then, $\tilde\Lc_\att^\st=\tilde\Lc_\pgd^\st$.
\end{proposition}

\begin{proof}
To begin with, let attention weights be
\begin{align*}
    \WQ\WK^\top=\begin{bmatrix}
        \Wb_1&\w_{1}\\
        *&*
    \end{bmatrix}\quad\text{and}\quad\WV=\begin{bmatrix}
        \Wb_2&\w_{2}\\
        \w_{3}^\top&w
    \end{bmatrix},
\end{align*}
where $\Wb_{1,2}\in\R^{d\times d},\w_{1,2,3}\in\R^{d}$ and $w\in\R$. Additionally, let task-specific prompts and heads be
\begin{align*}
    \pb_k=\begin{bmatrix}
    \pbb_k\\ p_k
    \end{bmatrix}\quad\text{and}\quad\hb_k=\begin{bmatrix}
    \hbb_k\\ h_k
    \end{bmatrix}\quad\text{for }k\in[K],
\end{align*}
where $\pbb_k,\hbb_k\in\R^{d}$ and $p_k,h_k\in\R$. Recapping the prediction from \eqref{eqn:LinearAttn_task} in the main paper and input sequence $\Zk$ from \eqref{def Z k} in the main paper, we obtain 
\begin{align*}
\tilde f_{\att}(\Zk)&=(\z^\top\WQ\WK^\top(\Zk)^\top)\M\Zk\WV\hb_k\\
&=\begin{bmatrix}
    \x^\top\Wb_1&\x^\top\w_1
\end{bmatrix}\left(\begin{bmatrix}\pbb_k\pbb_k^\top&p_k\pbb_k\\p_k\pbb_k^\top&p_k^2\end{bmatrix}+\begin{bmatrix}\X^\top\X&\X^\top\y\\\y^\top\X&\tn{\y}^2\end{bmatrix}\right)\begin{bmatrix}
    \Wb_2\hbb_k+h_k\w_2\\
    \w_3^\top\hbb_k+h_kw
\end{bmatrix}.
\end{align*}
Recap the task distribution from Definition~\ref{definition:multi-task}. Let $\y_0=\y-\X\bmu_k$. For cleaner notation and without loss of generality, we remove the subscription $k$ and set $$\begin{bmatrix}
    \Wb_2\hbb_k+h_k\w_2\\
    \w_3^\top\hbb_k+h_kw
\end{bmatrix}=\begin{bmatrix}
    \vb\\v
\end{bmatrix}$$
where $\vb\in\R^d$ and $v\in\R$. 
\begin{align*}
\tilde f_{\att}(\Zk)
&=\begin{bmatrix}
    \x^\top\Wb_1&\x^\top\w_1
\end{bmatrix}\left(\begin{bmatrix}\pbb\pbb^\top&p\pbb\\p\pbb^\top&p^2\end{bmatrix}+\begin{bmatrix}\X^\top\X&\X^\top\y\\\y^\top\X&\tn{\y}^2\end{bmatrix}\right)\begin{bmatrix}
    \vb\\v
\end{bmatrix}\\
&=\begin{bmatrix}
    \x^\top\Wb_1&\x^\top\w_1
\end{bmatrix}\begin{bmatrix}\pbb\pbb^\top&p\pbb\\p\pbb^\top&p^2\end{bmatrix}\begin{bmatrix}
    \vb\\v
\end{bmatrix}+\begin{bmatrix}
    \x^\top\Wb_1&\x^\top\w_1
\end{bmatrix}\begin{bmatrix}\X^\top\X&\X^\top\y\\\y^\top\X&\tn{\y}^2\end{bmatrix}\begin{bmatrix}
    \vb\\v
\end{bmatrix}\\
&=\scalemath{0.9}{\x^\top\left(\Wb_1\pbb\pbb^\top\vb+p\w_1\pbb^\top\vb+pv\Wb_1\pbb +p^2 v\w_1 \right)+\x^\top\left(\Wb_1\X^\top\X\vb+\w_1\y^\top\X\vb+v\Wb_1\X^\top\y +v\tn{\y}^2 \w_1 \right)}\\
&=\x^\top\tilde\pb+\x^\top\left(\Wb_1\X^\top\X\vb+(\w_1\vb^\top+v\Wb_1)\X^\top\y +v\tn{\y}^2 \w_1 \right)\\
&=\x^\top\left(\tilde\pb+\Wb_1\X^\top\X\vb+(\w_1\vb^\top+v\W_1)\X^\top(\y_0+\X\bmu) +v \w_1(\tn{\y_0}^2+\bmu^\top\X^\top\X\bmu+2\bmu^\top\X^\top\y_0) \right)\\
&=\x^\top\Wt\X^\top\y_0 +\x^\top\tilde\pb+\x^\top\underset{\eps(\X,\y_0)}{\underbrace{\left(\Wb_1\X^\top\X\vb+(\Wt-v \w_1\bmu^\top)\X^\top\X\bmu +v \w_1\tn{\y_0}^2\right)}}
\end{align*}
where
\begin{align*}
&\tilde\pb:=\Wb_1\pbb\pbb^\top\vb+p\w_1\pbb^\top\vb+pv\Wb_1\pbb +p^2 v\w_1\\
&\Wt:=\w_1\vb^\top+v\W_1+2v \w_1\bmu^\top.
\end{align*}
Then letting $y_0=y-\x^\top\bmu_k$, the expected risk of task $k$ obeys
\begin{align*}
    \E_{\Z,y\sim\Dc_k}[(\tilde f_\att(\Zk)-y)^2]&=\E\left[\left(\x^\top\Wt\X^\top\y_0-y_0 +\x^\top\tilde\pb-\x^\top\bmu+\x^\top\eps(\X,\y_0)\right)^2\right]\\
    &=\E\left[\left(\x^\top\Wt\X^\top\y_0-y_0 \right)^2\right]+\E\left[\left(\x^\top\tilde\pb-\x^\top\bmu+\x^\top\eps(\X,\y_0)\right)^2\right]\\
    &\quad+2\E\left[\left(\x^\top\Wt\X^\top\y_0-y_0\right)\left(\x^\top\tilde\pb-\x^\top\bmu+\x^\top\eps(\X,\y_0)\right)\right]
\end{align*}
Note that, letting $\bt_0=\bt-\bmu_k$, we have $\y_0=\X\bt_0+\mtx{\xi}$, $y=\x^\top\bt_0+\xi_{n+1}$ and $\bt_0\sim\Nc(0,\bSi_{\bt_k})$. Therefore, 
\begin{align*}
    &\E\left[\left(\x^\top\Wt\X^\top\y_0-y_0\right)\left(\x^\top\tilde\pb-\x^\top\bmu+\x^\top\eps(\X,\y_0)\right)\right]\\
    &=\E\left[\x^\top\left(\Wt\X^\top\X\bt_0-\bt_0\right)\left(\tilde\pb-\bmu+\eps(\X,\y_0)\right)^\top\x\right]\\
    &=\E\left[\x^\top\left(\Wt\X^\top\X-\Iden\right)\bt_0~\eps(\X,\y_0)^\top\x\right]\\
    &=\E\left[\x^\top\left(\Wt\X^\top\X-\Iden\right)\bt_0\left(v\tn{\X\bt_0}^2\right)\w_1^\top\x\right]\\
    &=0.
\end{align*}
Then the risk satisfies 
\begin{align*}
    \E_{\Z,y\sim\Dc_k}[(\tilde f_\att(\Zk)-y)^2]
    &=\E\left[\left(\x^\top\Wt\X^\top\y_0-y_0 \right)^2\right]+\E\left[\left(\x^\top\tilde\pb-\x^\top\bmu+\x^\top\eps(\X,\y_0)\right)^2\right]\\
    &\geq\E\left[\left(\x^\top\Wt\X^\top\y_0-y_0 \right)^2\right].
\end{align*}
We next prove that the equality is achievable for any $\Wt$. Consider the following constructions:
\begin{align*}
    \WQ\WK^\top=\begin{bmatrix}
        \W&\zerbb\\
        \zerbb^\top&0
    \end{bmatrix},\quad\WV=\Iden,\quad
    \pb_k=\begin{bmatrix}
    \W^{-1}\bmu_k\\ \bmu_k^\top\W^{-\top}\bmu_k+1
    \end{bmatrix}\quad\text{and}\quad\hb_k=\begin{bmatrix}
    -\bmu_k\\ 1
    \end{bmatrix}.
\end{align*}
Then
\[
\begin{bmatrix}
    \vb\\v
\end{bmatrix}=\begin{bmatrix}
    -\bmu_k\\1
\end{bmatrix},\quad\tilde\pb=\bmu_k,\quad\text{and}\quad\Wt=\W.
\]
Using above construction, we obtain that for any $\W\in\R^{d\times d}$, there exist $\pb_k$'s and $\hb_k$'s such that 
\[
\E_{\Z,y\sim\Dc_k}\left[\left(\x^\top\tilde\pb-\x^\top\bmu+\x^\top\eps(\X,\y_0)\right)^2\right]=0
\]
and hence,
\[
\E_{\Z,y\sim\Dc_k}[(\tilde f_\att(\Zk)-y)^2]=\E\left[\left(\x^\top\W\X^\top\y_0-y_0 \right)^2\right].
\]

Next, consider the preconditioned gradient descent problem defined in \eqref{eqn:obj pgd h} in the main paper. Recapping the PGD prediction where we have
\begin{align*}
    \tilde f_\pgd(\Zk)=\x^\top\W\X^\top(\y-\X\bmu_k)+\x^\top\bmu_k.
\end{align*}
Then
\begin{align*}
    \E_{\Z,y\sim\Dc_k}[(\tilde f_\pgd(\Zk)-y)^2]&=\E_{\Z,y\sim\Dc_k}[(\x^\top\W\X^\top(\y-\X\bmu_k)+\x^\top\bmu_k-y)^2]\\
    &=\E_{\Z,y\sim\Dc_k}[(\x^\top\W\X^\top\y_0-y_0)^2].
\end{align*}
Combining the results together completes the proof.
\end{proof}

\subsection{Proof of Theorem~\ref{thm:fully_decoupled}}
\begin{theorem}
 Consider the multi-task ICL problem with dataset defined in Definition~\ref{definition:multi-task}. Let $\W_\pgd^\st:=\arg\min_{\W}\Lc(\tilde f_\pgd)$ following \eqref{eqn:obj pgd h} in the main paper. Define $\bSbbt$ in \eqref{eqn:bt bias} in the main paper and let $\Wb_\pgd^\st=\bSi_{\x}\W_\pgd^\st$. Then the solution $\Wb_\pgd^\st$ and optimal loss $\tilde\Lc^\st_\pgd$ (c.f.~\eqref{eqn:obj pgd h} in the main paper) satisfy 
\begin{tcolorbox}[colback=white!5!white,colframe=black!5!black,colback=white!5!white,
                  interior hidden,
                  arc=0pt,
                  boxrule=1pt,
                  boxsep=0pt,
                  left=5pt,
                  right=5pt,
                  top=5pt,]
\[
\scalemath{0.9}{\Wb_\pgd^\star=\debias{\bSbbt}\left((n+1)\debias{\bSbbt}+(\tr{\debias{\bSbbt}}+\sigma^2)\Iden\right)^{-1},}
\]
\[
\scalemath{0.9}{\tilde\Lc^\star_{\pgd} 
=\tr{\debias{\bSbbt}}
-n\tr{\Wb_\pgd^\star\debias{\bSbbt}}.}
\]
\end{tcolorbox}
\end{theorem}
\begin{proof}
From the proof of Proposition~\ref{prop h}, it can be seen that the optimal multi-task ICL learning performance of a 1-layer linear attention model can be rigorously calculated as: $\tilde\Lc_\att^\st = \tilde\Lc_\pgd^\st$. Moreover, in this case, with the help of task-specific heads $\hb_k$, the optimal loss $\tilde\Lc^\star_{\pgd}$ is equivalent to the optimal plain training loss in a \textbf{zero task mean} multi-task ICL setting. 

By applying Theorem~\ref{thm:pretrain} with all task means $\bmu_k = \zerbb$ for $k \in [K]$, Theorem~\ref{thm:fully_decoupled} can be proven.

\end{proof}

%% file: app/Appendix/noisy.tex
We conduct experiments on synthetic datasets to validate our theoretical assumptions and explore the behavior of single-layer linear attention models with various trainable parameters under different training settings.

\textbf{Experimental Setting.} We train single-layer attention models to solve $K$-task, $d$-dimensional linear regression ICL with noise level $\sigma^2=5$. For each context length $n$, an independent model is trained for $20,000$ iterations with a batch size of $8192$ using the Adam optimizer (learning rate $10^{-3}$). 

To ensure robustness, each training process is repeated $50$ times with independent initializations, and the minimal test risk among these trials is reported. Theoretical predictions in the plots are based on the theorems in Section~\ref{sec:main}, and all results are normalized by $\E[\|y\|^2]$.
\subsection{Noisy labels}
We validate our theoretical assumptions and predictions based on a noise-free setting. To test these assumptions in a noisy label setting, where $\sigma^2 > 0$, we repeat the experiments from the main paper under the noisy setting, using the same experimental configurations as in Figure 2(a) and Figure 2(c), to validate Assumption~\ref{assumption:PGD} and Theorems~\ref{thm:pretrain}, \ref{thm:pretrain-finetune}, \ref{thm:joint-training}, and \ref{thm:fully_decoupled}.

\subsection{Non-isotropic covariance}
We also validate our experiments under a non-isotropic covariance setting. At noise level $\sigma^2 = 5$, we repeat the experiments from the main paper under the noisy setting, using the same experimental configurations as in Figure 2(a) and Figure 2(d), except that the isotropic covariance multiplier $\Iden_{10}$ is replaced with a non-isotropic one:
\[
\Iden_{10} \longrightarrow \texttt{diag}\{0.9, 0.9^{-1}, \dots, 0.9^{-9}\}
\]
This is done to validate Assumption~\ref{assumption:PGD} and Theorems~\ref{thm:pretrain}, \ref{thm:pretrain-finetune}, \ref{thm:joint-training}, and \ref{thm:fully_decoupled}.
\input{AISTATS_figs/supp/supp_fig_addtional}

As seen from Figure A1 (a)(c), the performance of the reduced model derived from Assumption~\ref{assumption:PGD} aligns perfectly with the performance of the exact linear attention model, indicating that it serves as a good proxy. Furthermore, from Figure A1 (b)(d), our theoretical predictions align perfectly with the performance of the linear attention model.

%% file: AISTATS_figs/supp/supp_fig_addtional.tex
\begin{figure*}[ht]
\vspace{-10pt}
\centering
~\hspace{-15pt}
\subfloat[]{\begin{tikzpicture}
\node at (-3.2,-.4)[anchor=west]{\includegraphics[width=0.25\textwidth]{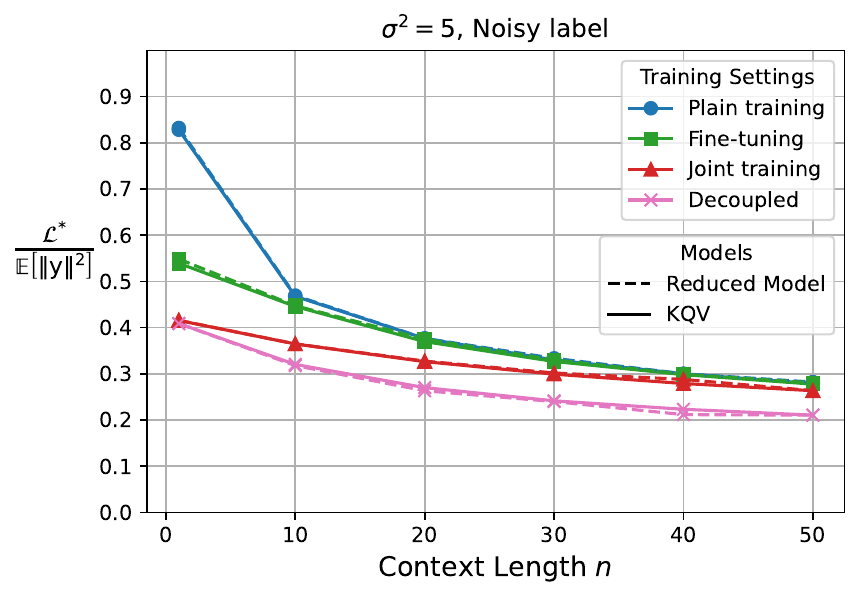}};
\node at (-1.1,1.35)[anchor=west][scale=0.6,opacity=1]{};
\node at (-2.9,-1.6)[anchor=west][scale=0.6,rotate=90,opacity=1]{ };
\end{tikzpicture}
}
~\hspace{-15pt}
\subfloat[]{\begin{tikzpicture}
\node at (-3.2,-.4)[anchor=west]{\includegraphics[width=0.25\textwidth]{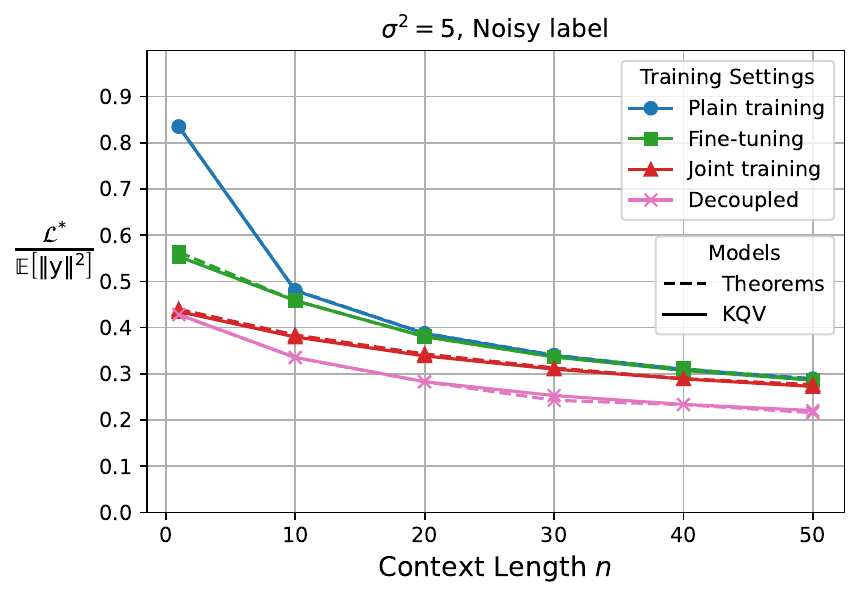}};
\node at (-0.7,1.35)[anchor=west][scale=0.6,opacity=1]{};
\end{tikzpicture}
}
~\hspace{-15pt}
\subfloat[]{\begin{tikzpicture}
\node at (-3.2,-.4)[anchor=west]{\includegraphics[width=0.25\textwidth]{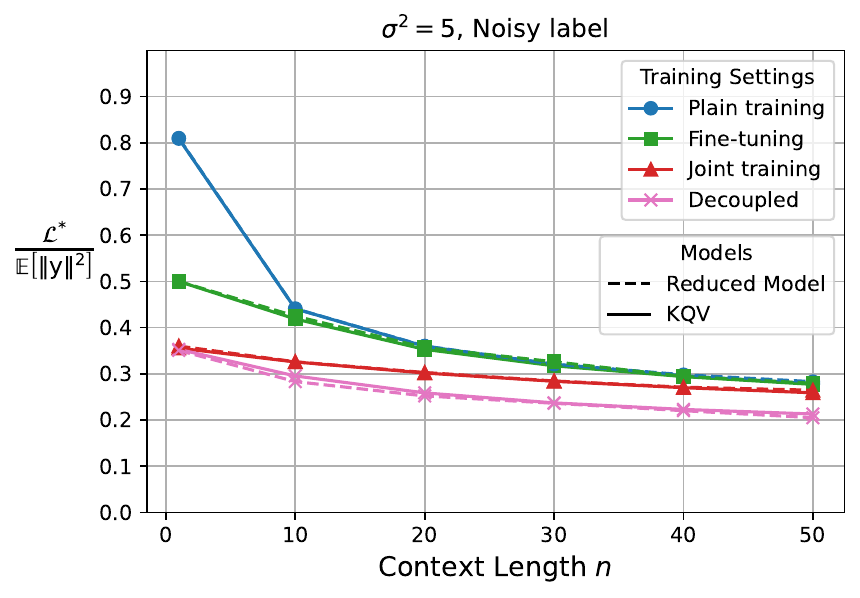}};
\node at (-1.1,1.35)[anchor=west][scale=0.6,opacity=1]{};
\node at (-2.9,-1.6)[anchor=west][scale=0.6,rotate=90,opacity=1]{ };
\end{tikzpicture}
}
~\hspace{-15pt}
\subfloat[]{\begin{tikzpicture}
\node at (-3.2,-.4)[anchor=west]{\includegraphics[width=0.25\textwidth]{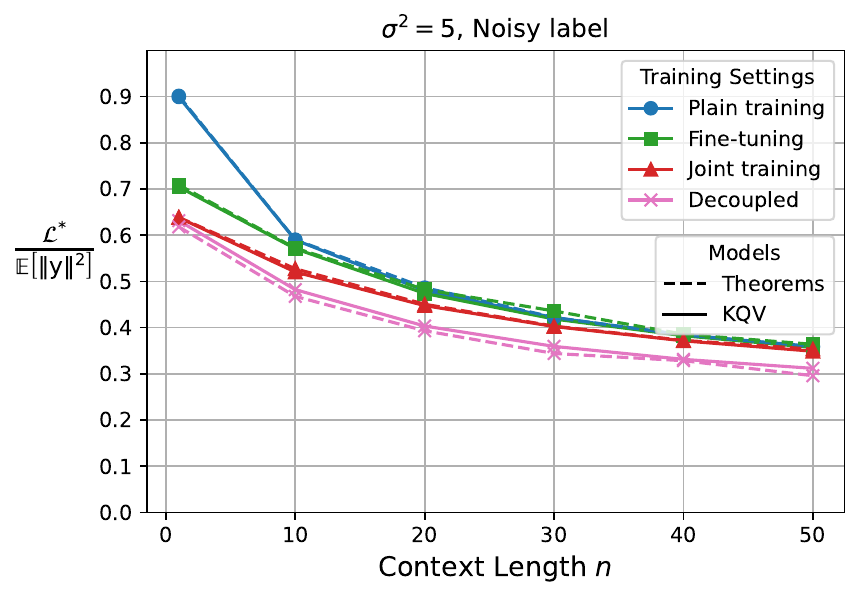}};
\node at (-0.7,1.35)[anchor=west][scale=0.6,opacity=1]{};
\end{tikzpicture}
}
\caption{Experimental results across various settings: (a)(b) Noisy labels; (c)(d) Non-isotropic covariance. (a)(c) validate Assumption 1 in the main paper, and (b)(d) validate Theorems 1-4 in the main paper.} 
\vspace{-0.17in}
\end{figure*}

%% file: app/Appendix/multilayer.tex
\subsection{Multi-layer linear attention model}
In the main paper, Section~\ref{sec:assumptions}, the output of a single-layer linear attention model is defined as:
\[
\att(\Zk) = (\Zk \WQ \WK^\top (\Zk)^\top)\M\Zk\WV.
\]
If task-specific heads $\hb_k, k \in [K]$ (see Section~\ref{sec:fully-decoupled}) are used, the prediction for $\Zk$ is (where $\mtx{e}_{i}$ is a one-hot indicator vector with 1 at the $i$-th position and 0 elsewhere):
\[
\hat{y} = \mtx{e}_{n+1}^\top \att(\Zk) \hb_k,
\]
otherwise,
\[
\hat{y} = \mtx{e}_{n+1}^\top \att(\Zk) \mtx{e}_{d+1}.
\]
Next, we extend the network architecture we used throughout this section. For a model with $L\ge2$ layers, we define an \textbf{$L$-layer linear attention model} as a stack of $L$ single-layer attention models. Formally, denoting by $\Zk_l$ the output of the $l$-th layer of attention, we define
\[
\Zk_{l+1} = \Zk_l + \att(\Zk_l), \quad l = 1, \dots, L-1.
\]
If task-specific heads $\hb_k, k \in [K]$ (see Section~\ref{sec:fully-decoupled}) are used, the prediction for $\Zk$ is :
\[
\hat{y} = \mtx{e}_{n+1}^\top \Zk_L \hb_k,
\]
otherwise,
\[
\hat{y} = \mtx{e}_{n+1}^\top \Zk_L \mtx{e}_{d+1}.
\]
\subsection{Experiments}
We only conduct empirical study to explore the impact of \textbf{linear attention model depth} and \textbf{label noise} in this section. Similarly, We conduct experiments on synthetic datasets to explore the behavior of multi-layer linear attention models with various trainable parameters under a joint training setting.

\input{AISTATS_figs/supp/supp_fig_multilayer}

\textbf{Experimental Setting.} For each context length $n$, an independent model is trained for $20,000$ iterations with a batch size of $8192$ using the Adam optimizer (learning rate $10^{-3}$). To ensure robustness, each training process is repeated $50$ times with independent initializations, and the minimal test risk among these trials is reported. All the results are normalized by $\E[\|y\|^2]$.

We use a same synthetic dataset configuration across all the multi-layer attention model experiments:
\[
\begin{aligned}
&\sigma^2=5 \text{ (Noisy), or } \sigma^2=0 \text{ (Noise-free)}\\
&d=10,\quad K=2,\quad \bSi_{\text{non-iso}}=\texttt{diag}\{0.9, 0.9^{-1}, \dots, 0.9^{-9}\}\\
&\M_\mu = \begin{bmatrix} 
    1.7 \cdot \onebb_{10} & 
    -1.3 \cdot \onebb_{10} 
    \end{bmatrix}, \\
    &\bSi_{\bt_1} = \frac{1}{2}\bSi_{\bt_2} = \bSi_{\text{non-iso}}, \quad \pi_1 = 0.3, \quad\pi_2 = 0.7
\end{aligned}
\]

As seen in Figure A2: (1) For a clean dataset, task-specific parameters significantly improve performance in the few-shot context region (where the context length $n < d$), but this benefit diminishes as $n$ increases. (2) Task-specific parameters help mitigate the impact of label noise. (3) Although increasing the depth of the attention model can narrow the performance gaps between different numbers of trainable task-specific parameters, adding task-specific parameters remains beneficial.

%% file: AISTATS_figs/supp/supp_fig_multilayer.tex
\begin{figure*}[ht]
\vspace{-10pt}
\centering

~\hspace{-15pt}
\subfloat[]{\begin{tikzpicture}
\node at (-3.2,-.4)[anchor=west]{\includegraphics[width=0.25\textwidth]{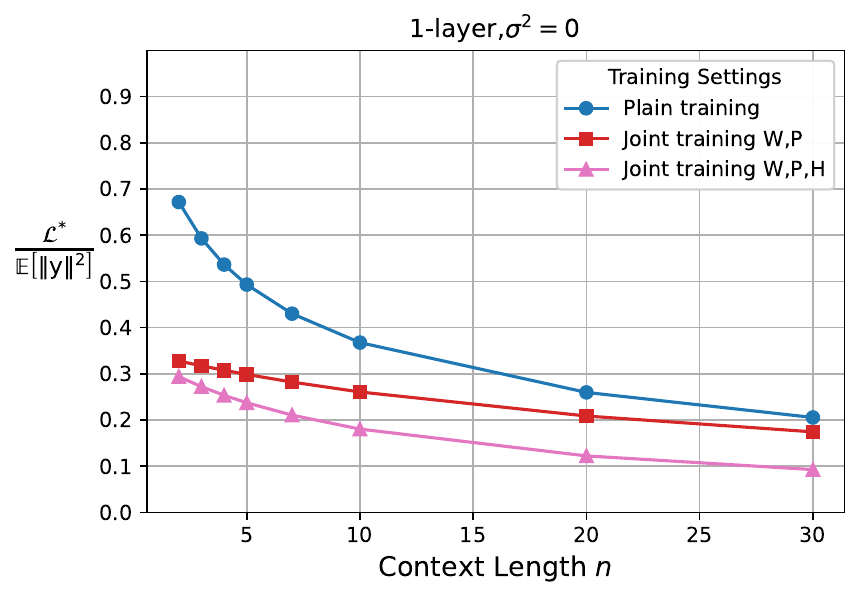}};
\node at (-1.1,1.35)[anchor=west][scale=0.6,opacity=1]{};
\node at (-2.9,-1.6)[anchor=west][scale=0.6,rotate=90,opacity=1]{ };
\end{tikzpicture}
}
~\hspace{-15pt}
\subfloat[]{\begin{tikzpicture}
\node at (-3.2,-.4)[anchor=west]{\includegraphics[width=0.25\textwidth]{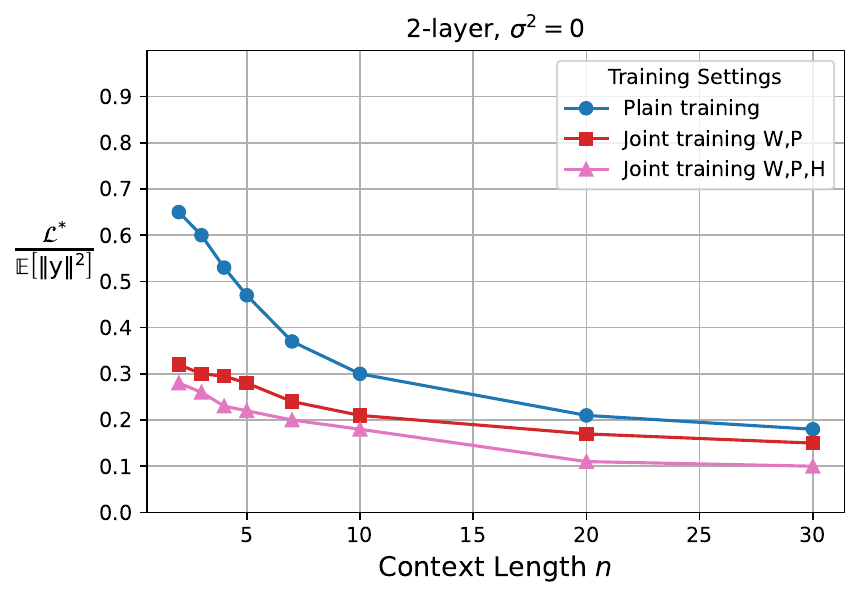}};
\node at (-0.7,1.35)[anchor=west][scale=0.6,opacity=1]{};
\end{tikzpicture}
}
~\hspace{-15pt}
\subfloat[]{\begin{tikzpicture}
\node at (-3.2,-.4)[anchor=west]{\includegraphics[width=0.25\textwidth]{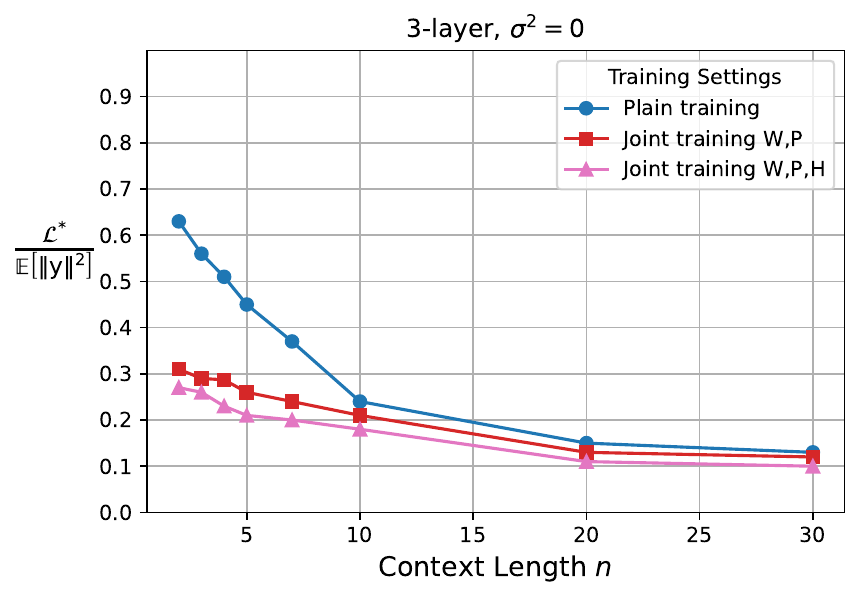}};
\node at (-1.1,1.35)[anchor=west][scale=0.6,opacity=1]{};
\node at (-2.9,-1.6)[anchor=west][scale=0.6,rotate=90,opacity=1]{ };
\end{tikzpicture}
}

\vspace{-10pt} %

~\hspace{-15pt}
\subfloat[]{\begin{tikzpicture}
\node at (-3.2,-.4)[anchor=west]{\includegraphics[width=0.25\textwidth]{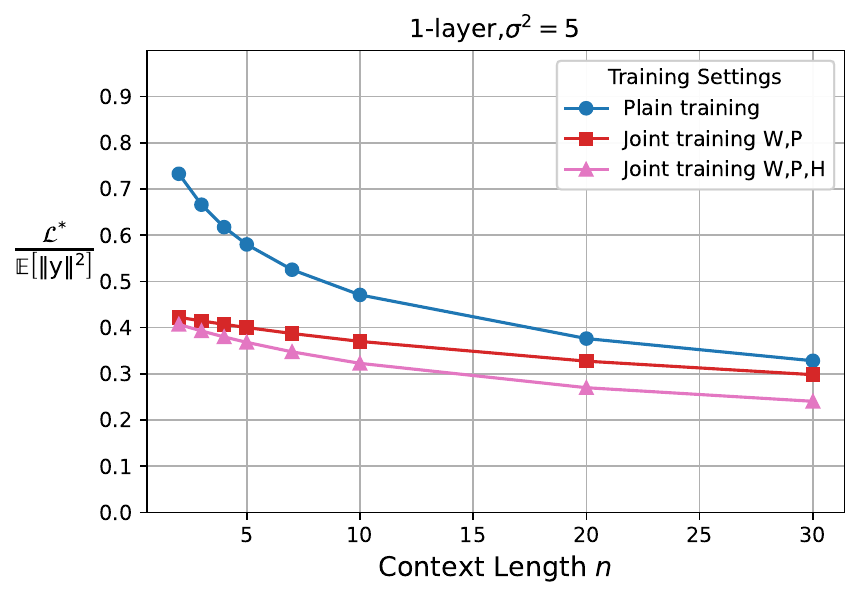
}};
\node at (-0.7,1.35)[anchor=west][scale=0.6,opacity=1]{};
\end{tikzpicture}
}
~\hspace{-15pt}
\subfloat[]{\begin{tikzpicture}
\node at (-3.2,-.4)[anchor=west]{\includegraphics[width=0.25\textwidth]{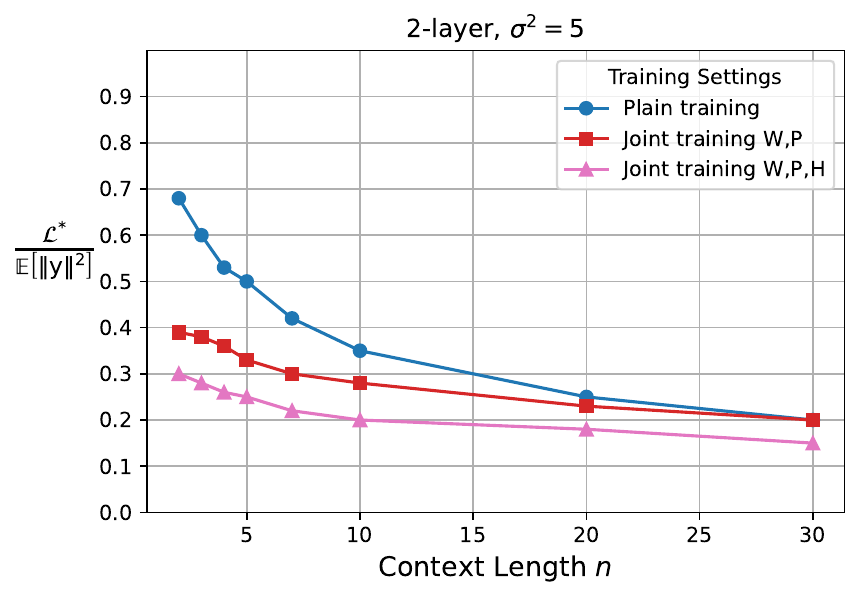}};
\node at (-1.1,1.35)[anchor=west][scale=0.6,opacity=1]{};
\node at (-2.9,-1.6)[anchor=west][scale=0.6,rotate=90,opacity=1]{ };
\end{tikzpicture}
}
~\hspace{-15pt}
\subfloat[]{\begin{tikzpicture}
\node at (-3.2,-.4)[anchor=west]{\includegraphics[width=0.25\textwidth]{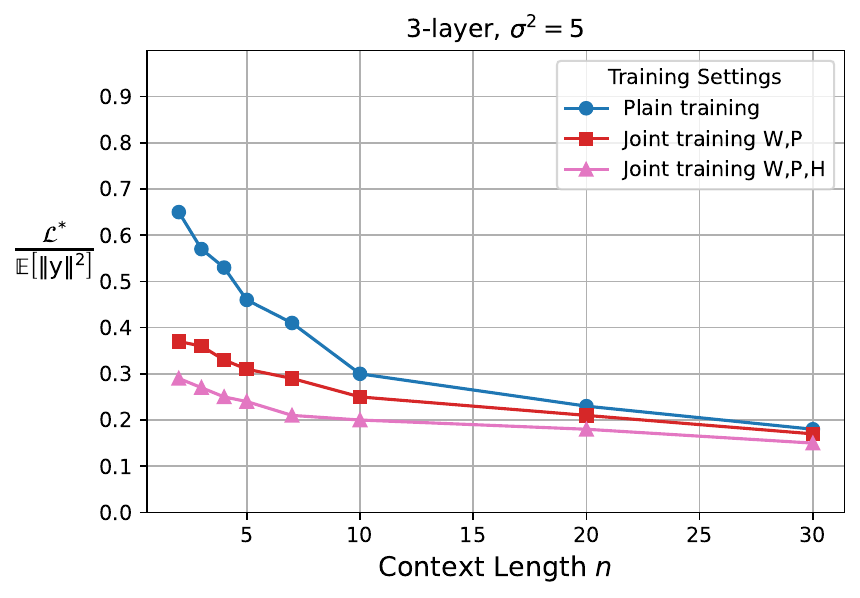}};
\node at (-0.7,1.35)[anchor=west][scale=0.6,opacity=1]{};
\end{tikzpicture}
}

\caption{Performance of $L$-layer linear attention models ($L=1, 2, 3$) on (a-c) clean and (d-f) noisy datasets.}
\vspace{-0.17in}
\end{figure*}